\documentclass[11pt]{article}

\usepackage[preprint]{acl}

\usepackage{times}
\usepackage{latexsym}

\usepackage[T1]{fontenc}

\usepackage[utf8]{inputenc}
\usepackage{float}
\usepackage{stfloats}

\usepackage{graphicx}
\usepackage{caption}
\usepackage{cuted}
\usepackage{microtype}
\usepackage{array}
\usepackage{amssymb}
\usepackage{inconsolata}

\usepackage{xcolor}
\usepackage{listings}
\usepackage{booktabs}
\usepackage{multirow}
\usepackage{framed}
\colorlet{shadecolor}{gray!8}

\usepackage{amsmath}
\title{FAM-Bench: Benchmarking Condition-Aware Food-as-Medicine Reasoning in Vision-Language Models}

\author{
  \textbf{Mingyang Mao}\textsuperscript{1,*},
  \textbf{Bhargav Rishi Medisetti}\textsuperscript{2,*},
  \textbf{Utkarsh Grover}\textsuperscript{1,*},
  \textbf{Tanvir Ibrahim}\textsuperscript{2},
\\
  \textbf{Wenyan Li}\textsuperscript{3},
  \textbf{Tingting Zhang}\textsuperscript{2},
  \textbf{Xiaomin Lin}\textsuperscript{1,\textdagger}
\\
\\
  \textsuperscript{1}Department of Electrical Engineering, University of South Florida
\\
  \textsuperscript{2}Muma College of Business, University of South Florida
\\
  \textsuperscript{3}Computer Science, University of Copenhagen
\\
\\
  \textsuperscript{*}Equal contribution.
\\
  \textsuperscript{\textdagger}Corresponding author: \text{xlin2@usf.edu}
}

\begin{document}
\maketitle




\begin{abstract}
Food AI can recognize dishes, ingredients, and nutrients. However, these
abilities do not show whether a dish is suitable for a person with a
specific health condition. They also do not test whether such decisions
are grounded in the relevant food evidence. To address this gap, we introduce FAM-Bench, a Food-as-Medicine benchmark comprising 2,500 question instances constructed from 1,811 recipes across 13 diet-related health conditions. Gold annotations are derived from rule-based dietary guidance informed by published sources and nutrition professionals. The benchmark contains two complementary tasks: dish-level suitability assessment 
and comparative dish analysis. 
We evaluate open-weight and proprietary vision-language models under four prompting modes. Results reveal a substantial gap between suitability decisions and their supporting rationales, as well as limited performance on comparative ranking. These findings highlight persistent challenges in grounded, condition-aware food reasoning.


\end{abstract}
\section{Introduction}

\vspace{0.5em}
\noindent\emph{``Let food be thy medicine.''}\hfill -- Hippocrates
\vspace{1em}

Diet is a major modifiable factor in chronic disease prevention and management. Dietary patterns are strongly associated with cardiometabolic and gastrointestinal conditions, including cardiovascular disease, diabetes, obesity, hypertension, and related disorders~\citep{willett1994diet,mozaffarian2016dietary}. Chronic diseases also impose substantial clinical and economic burden, accounting for nearly 90\% of the United States' \$4.9 trillion annual health-care expenditure~\citep{CDC2025ChronicDisease}. These pressures have renewed interest in \emph{Food As Medicine}, which integrates clinically appropriate food resources into health care to prevent, manage, or treat disease~\citep{Volpp2023FIM}.



\begin{figure*}[t]
    \centering
    \includegraphics[width=\textwidth]{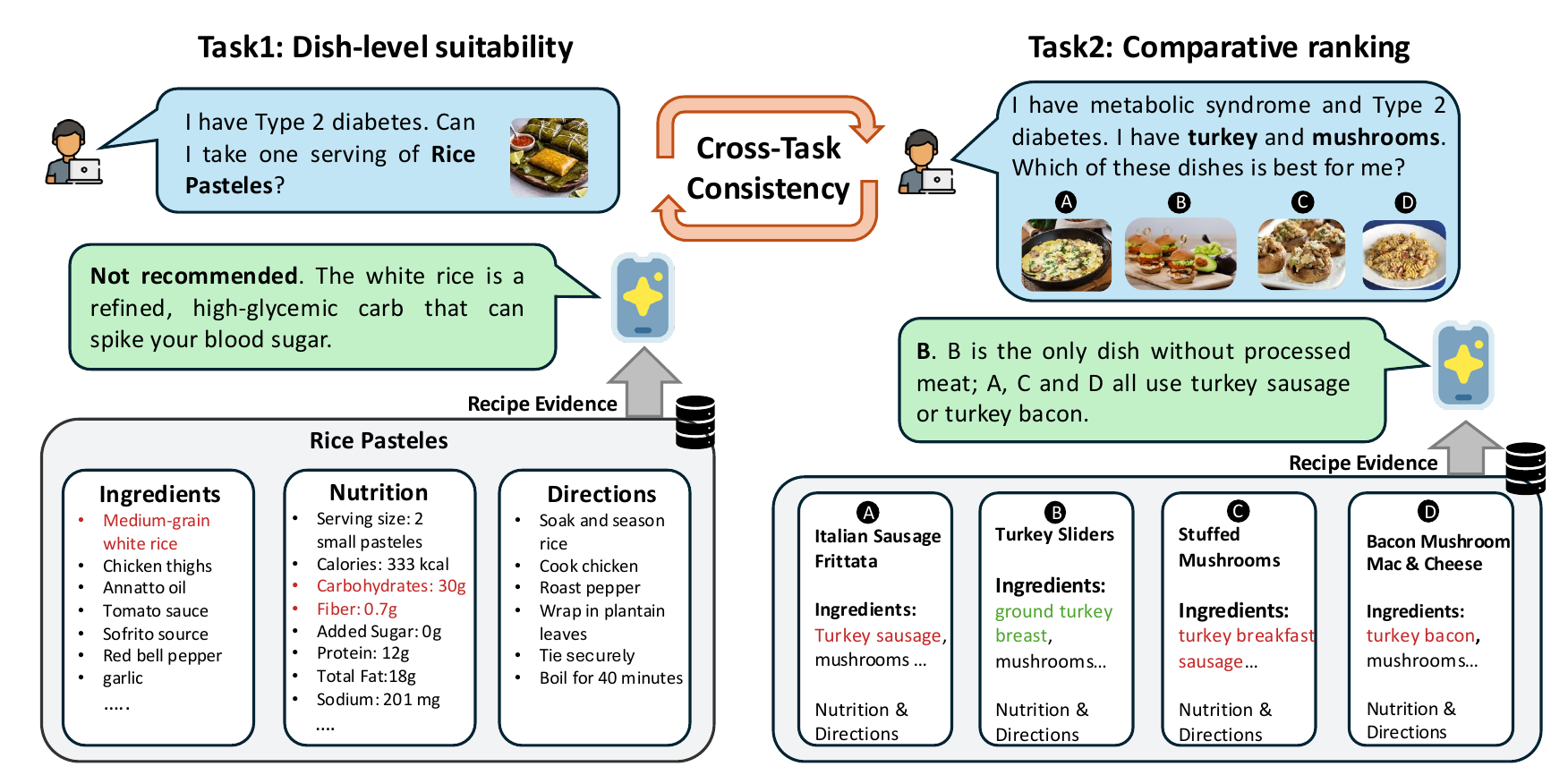}
\caption{\textbf{The two FAM-Bench tasks.}
\textbf{Task 1} (dish-level suitability): given a recipe and the user's
condition, the model labels the dish as recommended, neutral, or not
recommended and identifies the ingredients that determine the label
(here, white rice for type 2 diabetes).
\textbf{Task 2} (comparative ranking): given the user's conditions,
available ingredients, and four candidate recipes, the model ranks the
candidates (here, B is best because A, C, and D contain processed meat).
\textbf{Cross-task consistency} checks whether a model's rankings agree
with its own Task 1 labels for the same dishes.}
\label{fig:task_overview}
\vspace{-0.4cm}
\end{figure*}



For AI systems to support Food As Medicine, the central question is not what a dish is but whether it suits a specific health condition. Even a generally healthy dish can be the wrong choice for a particular condition: leafy greens, for example, may need to be limited in advanced chronic kidney disease because of their potassium. Existing food AI benchmarks mainly evaluate dish recognition, recipe understanding, ingredient extraction, nutrient estimation, or dietary visual question answering~\citep{bossard2014food,marin2021recipe1m+,yagcioglu2018recipeqa,wroblewska2022tasteset,thames2021nutrition5k,hua2024nutribench,jin2026diningbench}. These tasks provide important foundations but remain largely descriptive. NGQA~\citep{zhang2025ngqa} goes further by evaluating food suitability for individual users through reasoning over a structured graph of users, foods, nutrients, and health conditions. However, its inputs represent ingredients and nutritional properties through graph relations and predefined nutrient tags, rather than directly presenting recipe records and nutrition tables.


We introduce \textbf{FAM-Bench},\footnote{Code, data, and evaluation scripts are released at an anonymized repository for review: \url{https://github.com/anonymous-research-artifact123/Food-as-medicine}. The repository will be de-anonymized upon acceptance.} a Food-as-Medicine benchmark that grounds condition-specific dietary decisions in recipe evidence (Figure~\ref{fig:task_overview}). FAM-Bench contains 2,500 instances spanning 13 diet-related health conditions, built from 1,811 unique recipes, with labels derived by deterministic rules over curated clinical dietary guidance, including condition-specific dietary notes provided by professional nutritionists. The two tasks mirror two everyday decisions. In \emph{dish-level suitability assessment}, a user has a specific recipe in mind and asks whether it suits their conditions.
In \emph{comparative dish analysis}, a user has a few ingredients on hand and must choose among four candidate recipes.


We evaluate FAM-Bench on eight vision-language models: GPT-5.4~\citep{openai2025gpt54}, Claude Sonnet 4.6~\citep{anthropic2026claudesonnet46}, and six open-weight models from the Gemma~\citep{kamath2025gemma3,gemmateam2026gemma4technicalreport} and Qwen~\citep{bai2025qwen3vl,team2026qwen3_5} families, each under four prompting strategies: a direct baseline, chain-of-thought (CoT)~\citep{wei2022cot}, knowledge injection (KI), and CoT+KI. 
We further test whether each model's Task 2 rankings agree with its own Task 1 labels for the same dishes, evaluate photo-only and photo-augmented inputs, and probe perception directly by asking models to list the ingredients and cooking methods visible in each dish photo, with and without the dish title. We also collect a non-expert human baseline on a small subset of each task.

Our contributions can be summarized as: 

 \textbf{(I) Problem formulation.} We frame Food-as-Medicine
    evaluation as condition-specific dietary decisions grounded in
    recipe evidence, both for a single dish and among candidate dishes.
 \textbf{(II) Benchmark construction.} We release FAM-Bench, with
    2,500 instances from 1,811 recipes across 13 health conditions,
    labeled by traceable rules over clinical dietary guidance.
 \textbf{(III) Evaluation protocol.} We score decisions,
    ingredient-level rationales, rankings, and cross-task consistency,
    with additional photo-input and perception probes.
 \textbf{(IV) Empirical findings.} Across eight VLMs, the best
    open-weight models rival proprietary ones on dish-level assessment,
    yet models often miss the ingredients that drive a label, rank
    dishes against their own labels, and gain little from adding photos
    to recipes.

\section{Related Work}

\begin{figure*}[t!]
\vspace{-0.4cm}
\centering
\includegraphics[width=\linewidth]{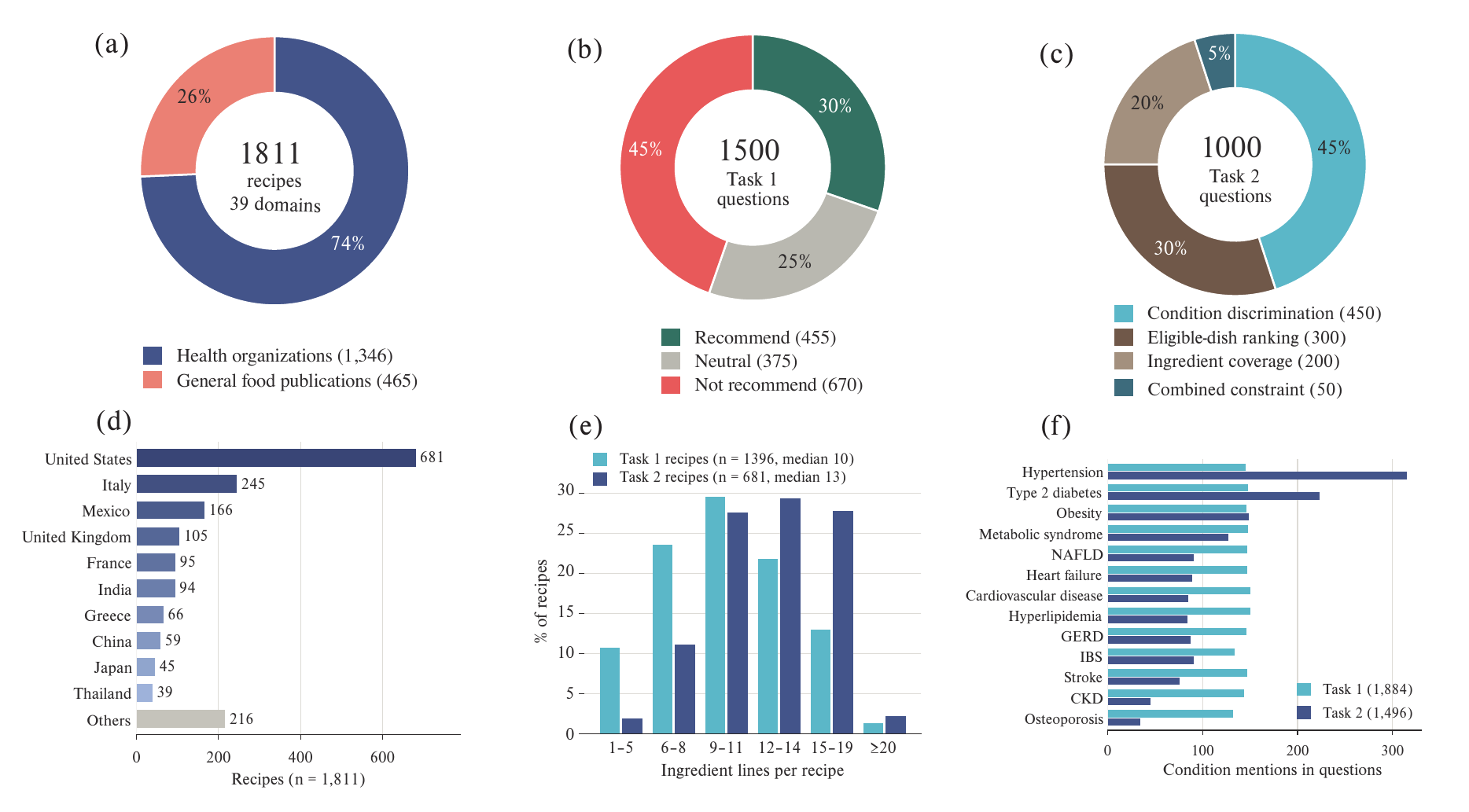}
\vspace{-0.6cm}
\caption{FAM-Bench statistics. (a) Sources of the 1{,}811 recipes.
(b) Task~1 gold labels. (c) Task~2 question types. (d) Recipe
cuisines, labeled by GPT-5.4 (top 10 shown). (e) Ingredient lines
per recipe. (f) Condition mentions per task; a question naming
several conditions counts once for each.}
\label{fig:benchmark_stats}
\vspace{-0.4cm}
\end{figure*}

\paragraph{Food understanding and nutrition benchmarks.}
Food AI benchmarks have largely focused on understanding \emph{what a food is}
and \emph{what it contains}. Food-101 evaluates dish classification
\cite{bossard2014food}, Recipe1M+ studies image recipe retrieval
\cite{marin2021recipe1m+}, RecipeQA tests multimodal procedural comprehension \cite{yagcioglu2018recipeqa}, and TASTEset targets structured recipe-entity extraction \cite{wroblewska2022tasteset}. Nutrition benchmarks move from
identity toward composition, estimating nutrients from dish images in Nutrition5k \cite{thames2021nutrition5k} or from meal descriptions in
NutriBench \cite{hua2024nutribench}; DiningBench further includes generic
dietary-suggestion questions in its VQA tier \cite{jin2026diningbench}.
Together, these benchmarks establish the perceptual and nutritional foundations needed for dietary reasoning, but largely stop before the decision layer: given a concrete dish and a particular health condition, \emph{should this person eat it?}

\paragraph{From food understanding to health-aware dietary decisions.}
A second line of work begins to address this decision layer through
personalized dietary guidance. Systems such as ChatDiet, NutriGen, and HealthGenie generate meal recommendations from user preferences, nutrition goals, and knowledge graphs \cite{yang2024chatdiet,khamesian2025nutrigen,gao2025healthgenie}, while
retrieval-augmented recommenders incorporate the medicine--food homology principle from traditional Chinese medicine \cite{gong2024integrating,sha2025leveraging}. FoodSky evaluates broad culinary and dietetic knowledge in a food-domain LLM \cite{zhou2025foodsky}. Most
closely related, NGQA asks whether a food is healthy for a specific user through graph question answering over NHANES/FNDDS nutrient records \cite{zhang2025ngqa}, extending earlier work that formulated food
recommendation as constrained QA over a food knowledge graph
\cite{chen2021personalized}. These works move from describing food toward health-conditioned judgment, but typically operate over structured nutrient records, knowledge graphs, or system-specific recommendation settings rather than asking models to judge real dishes directly from the evidence available to a user their recipe and photo.

\paragraph{Evaluating whether those decisions are grounded and reliable.}
Once models are asked to make health-conditioned dietary decisions, correctness alone is insufficient: the decision should also be supported by the evidence that justifies it. This connects FAM-Bench to the broader literature on medical-LLM evaluation, which has expanded beyond exam accuracy to potential harm \cite{singhal2023multimedqa}, hallucination \cite{pal2023medhalt}, and clinician-defined evaluation suites
\cite{bedi2025medhelm}. Chain-of-thought prompting and its medical variants \cite{wei2022cot,kojima2022zeroshot,nori2023medprompt}, together with
retrieval augmentation \cite{lewis2020rag,xiong2024medrag}, are common ways to improve knowledge-intensive medical reasoning. In the food domain, FoodGuardBench evaluates LLM safety on FDA-grounded food-safety questions, including adversarial settings \cite{luo2026cookingup}. These efforts motivate not only asking whether a model reaches the correct dietary judgment, but also whether that judgment is grounded in the condition-relevant ingredients and remains coherent when the same foods are compared against one another.

FAM-Bench brings these strands together by evaluating whether models can make condition-specific dietary decisions from real recipe and visual evidence, ground those decisions in the relevant ingredients, and remain consistent when moving from single-dish assessment to ranking alternatives. Appendix~\ref{app:related_work} provides a broader discussion of adjacent food, nutrition, and medical-LLM benchmarks.

\section{Benchmark Construction}
\label{sec:methodology}



FAM-Bench contains 2500 instances over 13 diet-sensitive conditions:
1500 for \emph{dish-level suitability} and 1000 for
\emph{comparative dish analysis} (\S\ref{sec:task-formulation}), built
from 1811 recipes from health organizations and food publications (Figure~\ref{fig:benchmark_stats}a) (\S\ref{sec:recipe-collection}). Gold labels follow deterministic
rules over nutritionist guidance, published food--condition mappings,
and nutrient limits, and are validated by nutritionists on a held-out
sample (\S\ref{sec:dataset-creation}). Figure~\ref{fig:benchmark_stats} shows the benchmark statistics.

\subsection{Recipe Collection}
\label{sec:recipe-collection}

\paragraph{Sources.}

The benchmark uses 1{,}811 recipes from 39 websites in two groups:
\emph{health organizations} (17 sites, 1{,}346 recipes), whose recipe
portals are run by medical societies, clinical nutrition programs, and
public health agencies, and \emph{food publications} (22 sites, 465
recipes), which include food magazines and cooking sites, some curated
by dietitians. Table \ref{tab:recipe-source-inventory} in Appendix~\ref{sec:appendix-recipe-sources} lists every
site. The dishes span 54 national cuisines, with the full distribution shown in Figure~\ref{fig:benchmark_stats}d.

\paragraph{Recipe preprocessing.}
We extract recipe content and metadata from source webpages using
\texttt{recipe-scrapers}, supplemented by site-specific parsing rules
where necessary. The extracted recipes are normalized into structured
records, with nutrition values expressed per serving and preparation
instructions standardized for consistent formatting. Processing is
largely automated, with manual intervention limited to parsing failures,
missing or inconsistent information, and ambiguous metadata.
User-generated metadata, including comments and ratings, are removed.
The benchmark recipes are drawn from a larger corpus, so that a recipe
can be replaced when its source page is unusable, for example when the
photo does not clearly show the dish or the ingredients cannot be
parsed. Figure~\ref{fig:benchmark_stats}e summarizes the distribution of ingredient-list lengths across the two tasks. The record schema is provided in
Appendix~\ref{sec:appendix-recipe-record}.

\subsection{Task Formulation}

\label{sec:task-formulation}

Each dish is \(d=(I,R)\), where \(I\) is the photo from its recipe
page and \(R\) is its recipe text: the title, ingredient list, plus the per-serving nutrition table. Each question names a
set \(H\subseteq\mathcal{H}\) of conditions, where \(\mathcal{H}\)
holds 13 diet-sensitive conditions spanning cardiometabolic,
gastrointestinal, hepatic, and renal disorders (Table~\ref{tab:condition_distribution}). On this representation we
define two complementary tasks that probe absolute and relative
dietary judgment, respectively. Example question layouts for the two tasks are shown in
Appendix~\ref{sec:appendix-example-questions}.

\paragraph{Dish-Level Suitability Assessment.}\label{sec:dish-level}
Given one dish \(d\) and one or two conditions \(H\), the model
answers ``Is this dish suitable for someone managing \(H\)?'' with one
of three labels, \textsc{recommend}, \textsc{neutral}, or
\textsc{not\_recommend}. 

A dish is \textsc{not\_recommend} if it conflicts with any condition in \(H\),
\textsc{recommend} if it benefits every condition in \(H\), and
\textsc{neutral} otherwise; Section~\ref{sec:dataset-creation} gives
the exact rule. For each condition, the model must also name the ingredients that
support its label. We compare them with the gold rationale
(Section~\ref{sec:dataset-creation}) to test whether a correct label
rests on the right evidence.

\paragraph{Comparative Dish Analysis.}\label{sec:comparative}
Given four dishes \(\mathcal{D}=\{d_1,\ldots,d_4\}\), one to three
co-occurring conditions \(H\) (e.g., hypertension and type~2
diabetes), and two to four ingredients \(U\) that the user has on
hand, the model ranks the dishes from most to least suitable. 
In the gold ranking \(\pi^\star\), dishes that contain no food to avoid
for \(H\) come first. Next, dishes that use all the available
ingredients rank higher, and a rule-based health score orders the rest
(Section~\ref{sec:dataset-creation}). This mirrors everyday decisions,
where people compare several dishes rather than judge one in
isolation. Because question types differ in what sets the gold dish
apart, such as a condition conflict or a missing ingredient, it is harder to rank
the dishes from the dish name or a single ingredient alone.

\subsection{Dataset Creation}

\label{sec:dataset-creation}

The 1811 dish records from Section~\ref{sec:recipe-collection} are
converted into the 2{,}500 benchmark instances of the two tasks
defined in Section~\ref{sec:task-formulation} through a rule-grounded annotation pipeline.

\paragraph{Knowledge base.}
Task~1 labels are derived from a knowledge base $\mathcal{K}$ that
lists, for each condition, beneficial foods and foods to limit or
avoid (with aliases for matching), together with per-serving nutrient
limits. The food lists combine dietary notes from nutritionists with
published food condition guidance (e.g., AHA, NIH, Harvard Nutrition
Source). Sodium must stay below its daily target, applied here to a
single serving: 2{,}300\,mg in general, 1{,}500\,mg for hypertension,
and 2{,}000\,mg for CKD. Added sugar may not exceed 10\,g. Saturated
fat fails a condition at a third of its daily value (6.7\,g) and caps
it at neutral once it reaches 10\% of energy.
Appendix~\ref{sec:appendix-knowledge-base} details the schema and
sources.



\paragraph{Dish-level questions.}
 GPT-5.5 screens each recipe against all 13 conditions, identifying potentially suitable conditions, $\mathrm{Rec}(d)$, and conflicting conditions, $\mathrm{NotRec}(d)$. We sample one condition from either set or two from the same set. Gold labels are determined independently using the knowledge base $\mathcal{K}$: foods to avoid and violated nutrient limits take precedence over beneficial foods, while a beneficial judgment requires matching foods and complete, compliant nutrition data. And the distribution is shown in Figure~\ref{fig:benchmark_stats}b. Gold rationales identify the foods driving each condition-level judgment. Appendix~\ref{sec:appendix-validation-rules} details the labeling rules and exceptions. The 1,500 questions cover each condition 132–150 times (Figure~\ref{fig:benchmark_stats}f).

\paragraph{Comparative questions.}

Task 2 questions were generated by our recipe recommender using separate condition rules (Appendix~\ref{sec:appendix-knowledge-base}). We sample from 23 condition sets and 32 common available ingredients. Each candidate undergoes two checks: whether its title, ingredients, and tags contain foods to avoid, and whether its ingredient list covers all available ingredients. The gold option passes both checks, while distractors are constructed to fail the condition check (450 questions), ingredient check (200), either or both (50), or neither (300), with the last category distinguished by a rule-based health score (Figure~\ref{fig:benchmark_stats}c). The 1,000 questions use 681 recipes, each appearing at most 15 times. Gold positions are balanced across A–D, and options are shuffled at evaluation time.


\paragraph{Expert guidance.}
Seven nutrition professionals reviewed our labeling rules:
three registered dietitians (RD/RDN), one nutrition and dietetics
technician, registered (NDTR), one reviewer with a degree in food
science, and two with degrees in health and behavioral science. Each
labeled ten questions without seeing the gold label and named the
ingredients or nutrients behind each judgment. Their feedback led to
three rule revisions: saturated fat is now graded in two tiers instead
of failing every dish at 10\% of energy; gaps in the food lists were
closed (e.g., vegetables for hypertension, white-rice varieties,
onions for GERD); and butter-like foods no longer veto dishes low in
saturated fat. More details are included in Appendix~\ref{sec:appendix-expert-review}.

\section{Experiments}

\subsection{Models}
\label{sec:models}


We evaluate eight instruction-tuned vision-language models: six
open-weight models (8B--32B parameters) served locally with
vLLM~\citep{kwon2023vllm}, namely Gemma-3-12B-IT and
Gemma-3-27B-IT~\citep{kamath2025gemma3}, Gemma-4-31B-IT,
Qwen3-VL-8B-Instruct and Qwen3-VL-32B-Instruct~\citep{bai2025qwen3vl},
and Qwen3.5-27B; and two proprietary models queried through their
official APIs, GPT-5.4~\citep{openai2025gpt54} and Claude Sonnet
4.6~\citep{anthropic2026claudesonnet46}. All models receive the same
prompts, decode at temperature~0, and run
with hidden reasoning (thinking) disabled, so explicit reasoning
appears only in the chain of thoughts (CoT)  modes. Per-model reasoning settings, output budgets, and serving details are in
Appendix~\ref{sec:appendix-inference}.

\subsection{Prompting Methods}
\label{sec:prompts}




All four prompting modes use the same recipe input and are scored the
same way. Since a three-way label is ambiguous on its own, every
Task~1 prompt defines the labels as in Section~\ref{sec:dish-level}:
a dish is \textsc{not\_recommend} if it conflicts with any named
condition, \textsc{recommend} only if it benefits all of them, and
\textsc{neutral} otherwise. The prompts never give numeric nutrient
limits, and only the Knowledge Injection (KI) modes tell the model which foods help or harm each condition. Otherwise, the model relies on its own knowledge.
Appendix~\ref{sec:appendix-prompts} lists all prompts verbatim.

\paragraph{Baseline.}
The model answers directly, relying on its own knowledge of which
foods suit each condition.


\paragraph{Chain-of-Thought (CoT).}
The model reasons step by step before answering, following a fixed
checklist rather than free-form CoT~\citep{wei2022cot}. In Task~1, it
names the conditions, looks for helpful and harmful ingredients,
checks the nutrition table, and applies the label definitions. In
Task~2, it checks each option for foods to avoid, beneficial foods,
and the available ingredients, then ranks the options by the same
priorities as the gold ranking (foods to avoid first, then ingredient
match). The reasoning is not scored.


\paragraph{Knowledge Injection (KI).}
For each named condition, the prompt adds a reference
$\mathcal{R}_h$: the beneficial and limit-or-avoid food lists of
$\mathcal{K}$, with no nutrient limits, labels, or rationales. Since
Tasks' labels come from these lists, KI tests whether a model can
apply the labeling knowledge once it is given; the model must still
match the recipe's ingredients to the lists and check the nutrition
table.

\paragraph{CoT + KI.}
The model receives $\mathcal{R}_h$ and follows the CoT checklist.

\subsection{Evaluation Metrics}
\label{sec:metrics}

\paragraph{Task 1: Dish-level suitability.}
\emph{Decision accuracy} is the fraction of instances whose
predicted three-way label matches the gold label, and \emph{decision
macro-F1} averages the F1 of the three labels. \emph{Rationale
macro-F1} and \emph{micro-F1} measure overlap between the
model-cited condition-relevant ingredients and the gold rationale for
the named conditions: macro-F1 averages per-instance F1 over
instances with a non-empty gold rationale; micro-F1 pools true
positives, false positives, and false negatives across all instances.


\paragraph{Task 2: Comparative ranking.}
\emph{Top-1 accuracy} is the fraction of queries whose top-ranked
dish matches the gold most-suitable dish in $\pi^\star$.
\emph{Mean Reciprocal Rank} (MRR) is
\begin{equation}
\mathrm{MRR} = \frac{1}{|Q|} \sum_{i=1}^{|Q|} \frac{1}{\mathrm{rank}_i},
\end{equation}
where $\mathrm{rank}_i$ is the position of the gold dish in the
returned ordering, taken as $\infty$ when absent. MRR awards partial
credit for near-miss rankings.


\paragraph{Cross-task consistency (CTC).}
For each query, we elicit pairwise orderings from the model's own
Task 1 per-option decisions under \textsc{rec} $>$ \textsc{neutral}
$>$ \textsc{not\_rec} and check whether its Task 2 ranking respects
them; pairs with equal decisions impose no order and are skipped. CTC
is the micro-averaged fraction of such pairs in agreement, with
chance $=50$. As both inputs come from the same model, CTC measures
internal coherence independent of ground truth.

\section{Results}

\subsection{Main Results}
\label{sec:main-results}

\begin{table*}[t]
\centering
\footnotesize
\setlength{\tabcolsep}{2.6pt}
\begin{tabular}{l >{\centering\arraybackslash}p{18pt} >{\centering\arraybackslash}p{18pt} >{\centering\arraybackslash}p{18pt} >{\centering\arraybackslash}p{18pt} @{\hspace{8pt}} >{\centering\arraybackslash}p{18pt} >{\centering\arraybackslash}p{18pt} >{\centering\arraybackslash}p{18pt} >{\centering\arraybackslash}p{18pt} @{\hspace{8pt}} >{\centering\arraybackslash}p{18pt} >{\centering\arraybackslash}p{18pt} >{\centering\arraybackslash}p{18pt} >{\centering\arraybackslash}p{18pt} @{\hspace{8pt}} >{\centering\arraybackslash}p{18pt} >{\centering\arraybackslash}p{18pt} >{\centering\arraybackslash}p{18pt} >{\centering\arraybackslash}p{18pt}}
\toprule
\textbf{Model} & \multicolumn{4}{c}{\textbf{Task 1 decision F1}} & \multicolumn{4}{c}{\textbf{Task 1 rationale F1}} & \multicolumn{4}{c}{\textbf{Task 2 top-1 acc.}} & \multicolumn{4}{c}{\textbf{Task 2 MRR}} \\
\cmidrule(lr){2-5} \cmidrule(lr){6-9} \cmidrule(lr){10-13} \cmidrule(lr){14-17}
\multicolumn{1}{r}{\scriptsize CoT} & -- & \checkmark & -- & \checkmark & -- & \checkmark & -- & \checkmark & -- & \checkmark & -- & \checkmark & -- & \checkmark & -- & \checkmark \\
\multicolumn{1}{r}{\scriptsize KI} & -- & -- & \checkmark & \checkmark & -- & -- & \checkmark & \checkmark & -- & -- & \checkmark & \checkmark & -- & -- & \checkmark & \checkmark \\
\midrule
\textit{Majority / random} & \multicolumn{4}{c}{20.7} & \multicolumn{4}{c}{--} & \multicolumn{4}{c}{25.0} & \multicolumn{4}{c}{52.1} \\
\midrule
Gemma-3-12B & 37.6 & 43.6 & \textbf{46.2} & 40.3 & 22.5 & 26.0 & \textbf{44.8} & 44.7 & \textbf{36.6} & 33.3 & 35.4 & 32.8 & \textbf{60.0} & 58.2 & 59.4 & 57.6 \\
Gemma-3-27B & 47.1 & 44.6 & \textbf{53.5} & 49.0 & 28.9 & 26.0 & \textbf{51.6} & 50.0 & 34.6 & 34.3 & \textbf{38.1} & 34.1 & 59.0 & 58.9 & \textbf{61.3} & 58.8 \\
Gemma-4-31B & 47.0 & 46.8 & \textbf{51.2} & 50.5 & 38.2 & 34.9 & 61.8 & \underline{\textbf{64.5}} & 34.5 & 35.8 & 38.5 & \textbf{39.4} & 58.7 & 58.2 & \textbf{61.2} & 60.4 \\
Qwen3-VL-8B & 44.0 & 38.7 & 49.9 & \textbf{51.9} & 20.9 & 8.6 & 42.6 & \textbf{52.8} & \textbf{31.4} & 29.4 & 30.9 & 29.8 & 56.7 & 53.1 & \textbf{56.9} & 54.5 \\
Qwen3-VL-32B & 44.9 & 47.0 & \textbf{53.0} & 51.1 & 22.1 & 31.3 & 49.5 & \textbf{56.6} & \textbf{37.4} & 32.5 & 35.8 & 33.9 & \textbf{60.7} & 56.9 & 59.9 & 58.1 \\
Qwen3.5-27B & 47.5 & 47.5 & \underline{\textbf{56.6}} & 51.5 & 32.7 & 36.6 & 55.9 & \textbf{60.4} & 34.7 & 36.4 & 33.7 & \textbf{41.8} & 59.0 & 59.5 & 58.7 & \textbf{63.0} \\
GPT-5.4 & 51.3 & 49.8 & \textbf{54.1} & 52.4 & 36.7 & 37.8 & 57.7 & \textbf{59.8} & 33.3 & 38.2 & 36.2 & \underline{\textbf{45.2}} & 58.1 & 61.2 & 60.6 & \underline{\textbf{65.0}} \\
Claude Sonnet 4.6 & 49.2 & 50.1 & 53.5 & \textbf{54.9} & 40.1 & 38.9 & 54.5 & \textbf{62.5} & 37.1 & 36.4 & 37.0 & \textbf{40.5} & 60.9 & 59.4 & 60.8 & \textbf{62.6} \\
\midrule
\textit{Mean} & 46.1 & 46.0 & \textbf{52.3} & 50.2 & 30.3 & 30.0 & 52.3 & \textbf{56.4} & 35.0 & 34.5 & 35.7 & \textbf{37.2} & 59.1 & 58.2 & 59.8 & \textbf{60.0} \\
\bottomrule
\end{tabular}
\caption{Main results with text-only input (all
values $\times 100$). \textbf{Task~1}: decision
macro-F1 over the three labels and rationale macro-F1 against the gold
ingredients. \textbf{Task~2}: top-1 accuracy and MRR. Within each metric, the four columns
are the prompting modes; unanswered or unparsable items count as
incorrect. Majority: always \textsc{not\_recommend}; random: uniform
ranking; \textit{Mean}: average over the eight models. \textbf{Bold}:
best mode per model and metric; \underline{underline}: best value per
metric.}
\label{tab:main_results}
\end{table*}


\begin{table}[h]
\centering
\small
\setlength{\tabcolsep}{5pt}
\begin{tabular}{l c c c c}
\toprule
\textbf{Model} & \textbf{Base} & \textbf{CoT} & \textbf{KI} & \textbf{CoT+KI} \\
\midrule
Gemma-3-12B & 54.3 & 57.4 & \textbf{60.3} & 58.9 \\
Gemma-3-27B & 64.6 & 62.5 & 67.9 & \textbf{70.8} \\
Gemma-4-31B & 76.5 & 77.9 & 82.9 & \underline{\textbf{91.5}} \\
Qwen3-VL-8B & 64.9 & 59.2 & \textbf{67.5} & 63.4 \\
Qwen3-VL-32B & 68.8 & 67.5 & 68.1 & \textbf{73.8} \\
Qwen3.5-27B & 71.3 & 72.8 & 80.4 & \textbf{82.1} \\
GPT-5.4 & 78.2 & 70.3 & 77.1 & \textbf{82.5} \\
Claude Sonnet 4.6 & 75.2 & 76.8 & 77.7 & \textbf{85.6} \\
\midrule
\textit{Mean} & 69.2 & 68.0 & 72.7 & \textbf{76.1} \\
\bottomrule
\end{tabular}
\caption{\textbf{Cross-task consistency (CTC)}, text-only input,
$\times 100$. Each Task~2 option is also judged alone with the Task~1
prompt; CTC is the share of option pairs, pooled over the 1{,}000
questions, whose Task~2 order agrees with these decisions
(\textsc{recommend} $>$ \textsc{neutral} $>$ \textsc{not\_recommend}).
Pairs with equal decisions are skipped. Random: 50; \textit{Mean}:
average over the eight models. \textbf{Bold}: best mode per model;
\underline{underline}: best value.}
\vspace{-0.4cm}
\label{tab:ctc}
\end{table}


\begin{table*}[t]
\centering
\small
\setlength{\tabcolsep}{4pt}
\begin{tabular}{l c c c c c c c c c c}
\toprule
 & \multicolumn{5}{c}{\textbf{Task 1 decision macro-F1}} & \multicolumn{5}{c}{\textbf{Task 2 top-1 accuracy}} \\
\cmidrule(lr){2-6} \cmidrule(lr){7-11}
\textbf{Model} & \textbf{P} & \textbf{R} & \textbf{R+P} & $\boldsymbol{\Delta_\text{photo}}$ & $\boldsymbol{\Delta_\text{recipe}}$ & \textbf{P} & \textbf{R} & \textbf{R+P} & $\boldsymbol{\Delta_\text{photo}}$ & $\boldsymbol{\Delta_\text{recipe}}$ \\
\midrule
Gemma-3-12B-IT & 39.2 & 41.6 & 42.0 & $+0.4$ & $+2.8$ & 31.5 & 34.5 & 30.9 & $-3.6$ & $-0.6$ \\
Gemma-3-27B-IT & 42.7 & 48.2 & 49.0 & $+0.8$ & $+6.3$ & 32.9 & 35.3 & 32.5 & $-2.8$ & $-0.4$ \\
Gemma-4-31B-IT & 43.2 & 48.6 & 48.1 & $-0.5$ & $+4.9$ & 34.6 & 37.0 & 37.4 & $+0.4$ & $+2.8$ \\
Qwen3-VL-8B-Instruct & 39.9 & 46.0 & 47.0 & $+1.0$ & $+7.1$ & 29.6 & 30.4 & 29.3 & $-1.1$ & $-0.3$ \\
Qwen3-VL-32B-Instruct & 42.7 & 48.9 & 48.7 & $-0.2$ & $+6.0$ & 33.0 & 34.9 & 33.9 & $-1.0$ & $+0.9$ \\
Qwen3.5-27B & 43.4 & 50.5 & 50.2 & $-0.3$ & $+6.8$ & 33.9 & 36.6 & 37.0 & $+0.4$ & $+3.1$ \\
\cmidrule(lr){1-11}
\textit{Open-weight mean} & 41.9 & 47.3 & 47.5 & $+0.2$ & $+5.6$ & 32.6 & 34.8 & 33.5 & $-1.3$ & $+0.9$ \\
\midrule
GPT-5.4 & 37.1 & 51.7 & 51.1 & $-0.6$ & $+14.0$ & 32.3 & 38.2 & 37.8 & $-0.4$ & $+5.5$ \\
Claude Sonnet 4.6 & 44.9 & 51.2 & 50.9 & $-0.3$ & $+6.0$ & 33.7 & 37.8 & 37.0 & $-0.8$ & $+3.3$ \\
\bottomrule
\end{tabular}
\caption{\textbf{Photo-only and photo-augmented inputs.}
Scores averaged over the four prompting modes.
\textbf{P}: dish title and photo (one per option in Task~2), no recipe
text; \textbf{R}: the recipe input of Table~\ref{tab:main_results};
\textbf{R+P}: both. $\Delta_\text{photo} = $ R+P $-$ R and
$\Delta_\text{recipe} = $ R+P $-$ P.}
\label{tab:input_settings}
\vspace{-0.2cm}
\end{table*}

Table~\ref{tab:main_results} reports both tasks under all four
prompting modes for the eight models, and Table~\ref{tab:ctc} their
cross-task consistency. We read the results through the
benchmark design: whether models can decide suitability, ground that
decision in ingredient evidence, compare alternatives, and remain
consistent across the two settings.

\paragraph{Decisions are easier than rationales.}
Models decide suitability far better than they identify the ingredients behind it. Averaged over the eight models, the baseline mode reaches a decision
macro-F1 of 46.1, far above the 20.7 of always predicting
\textsc{not\_recommend}, but a rationale macro-F1 of only 30.3. With
CoT, the mean rationale macro-F1 stays at 30.0. With KI, which gives the model the food lists behind
the labels, it rises to 52.3 (56.4 with CoT+KI). Knowing which foods matter for conditions is thus the main bottleneck.


\paragraph{Knowledge injection is more effective than CoT alone.}
The same holds for decisions: CoT leaves the mean decision macro-F1
unchanged (46.0 vs.\ 46.1), while KI raises it to 52.3. Adding CoT to
KI gives the best rationales (56.4) but lowers the decision macro-F1
to 50.2, below KI alone for six of the eight models, so no prompting
mode is best on every metric. By condition, decisions are weakest for
IBS, osteoporosis, CKD, and GERD, and KI halves their gap to the other
nine conditions (Appendix~\ref{sec:appendix-radar-modes})


\paragraph{Comparative ranking remains far from solved.}
Task~2 top-1 accuracy ranges from 29.4 to 45.2 across models and
modes, above the 25.0 of random ranking, and the mean rises only from
35.0 (baseline) to 37.2 (CoT+KI). Results differ by question type
(Table~\ref{tab:t2_types}). When the distractors violate a condition,
KI helps: mean top-1 rises from 36.0 to 40.9, and to 44.1 with CoT+KI.
When they lack an available ingredient, CoT and KI hurt (45.9 to 40.2
with KI), suggesting that health-focused prompts pull attention away
from the user's ingredients. In eligible-dish ranking, every model and
mode stays near chance (18.3--29.0): all four dishes pass both checks,
and the gold order rests on a health score computed from nutrition
values.

\paragraph{Consistency reveals failures not captured by accuracy.}
CTC checks whether a model ranks dishes in line with its own Task~1
decisions on the same dishes (Table~\ref{tab:ctc}). Every cell is
above the 50 of random ranking, but only Gemma-4-31B with CoT+KI
exceeds 90, and Gemma-3-12B (54.3--60.3) ranks almost independently of
its own decisions. KI makes the two tasks more consistent: mean CTC
rises from 69.2 (baseline) to 72.7 with KI and 76.1 with CoT+KI, and a
KI mode is the most consistent for every model. CoT alone does not
(68.0), and consistency need not follow accuracy: CoT raises GPT-5.4's
Task~2 top-1 from 33.3 to 38.2 but lowers its CTC from 78.2 to 70.3.


\paragraph{Non-expert human evaluation.}
Student volunteers answered 97 held-out Task~1 and 100 Task~2 questions
(Figure~\ref{fig:human_modality}). On Task~1 they reach $41.3$ decision
macro-F1, above the majority class ($20.7$) but below VLMs given the recipe
($48.6$), and they recommend $26.5\%$ of \textit{not recommend} dishes,
about twice the rate of baseline VLMs ($14.2\%$). On Task~2 they reach
$37.0\%$ top-1 (random $25\%$), on par with VLMs.

\subsection{Photo-Only and Photo-Augmented Inputs}
\label{sec:Photo-Only and Photo-Augmented Inputs}

We rerun the prompts of Table~\ref{tab:main_results} with two further
inputs (Table~\ref{tab:input_settings}): \emph{photo only} (P: dish
title and photo, no recipe text) and \emph{recipe plus photo} (R+P).
Since P and the recipe (R) carry different information, we compare
each only with R+P.

\paragraph{The photo adds little, and sometimes hurts.}
For the open-weight models, adding the photo to the recipe barely
moves Task~1 macro-F1 ($+0.2$) and lowers Task~2 top-1 ($-1.3$); the
closed models behave the same way. Adding the recipe to the photo
instead raises Task~1 macro-F1 by $5.6$, but Task~2 top-1 by only
$0.9$, so much of the ranking signal comes from the dish titles that
every setting includes. We trace the small photo gain to two causes:
how models use the photo, and what the photo can show.

\begin{figure*}[ht]
    \centering
\includegraphics[width=\linewidth]{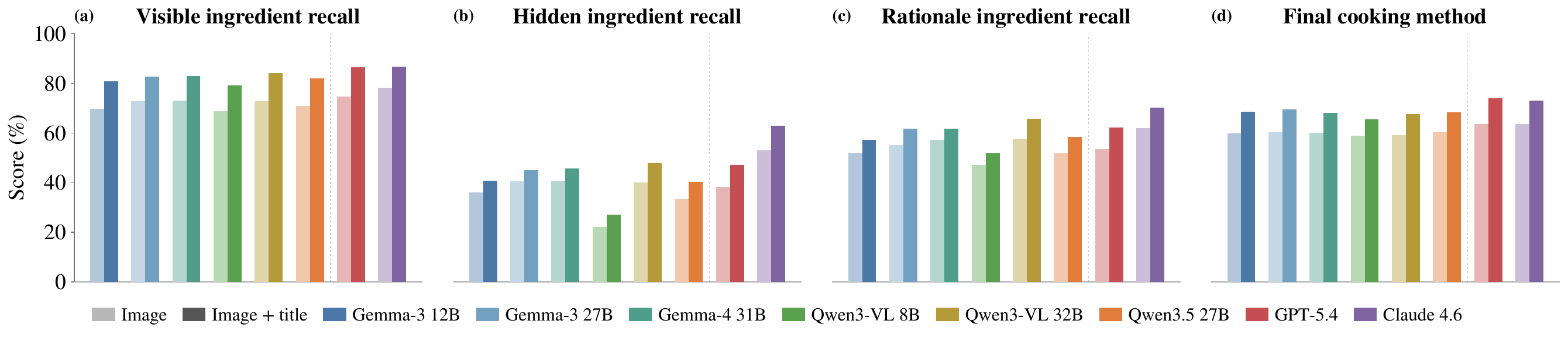}
\caption{\textbf{What models read from the dish photo.} Each model
lists the ingredients and cooking method of every Task~1 dish from its
photo, with or without the dish title. \textit{Hidden}: recipe
ingredients the photo does not show; dashed line in (d): majority
class. Per-model values in Table~\ref{tab:vision_probe}.}

    \label{fig:vision_probe}
    \vspace{-0.4cm}
\end{figure*}

\paragraph{The photo changes answers, not accuracy.}
A near-zero $\Delta_\text{photo}$ does not mean that models ignore the
photo. Compared question by question, the photo changes many answers
in both tasks, for open-weight and closed models alike, but it fixes
about as many as it breaks (Appendix~\ref{sec:appendix-photo-flips}).
It thus acts as a weakly grounded signal that perturbs text-based
decisions rather than correcting them. This agrees with
\citet{xu2026vision}, who hold the task information fixed across
text, image, and image+text inputs and find that adding the image
often lowers accuracy relative to text alone: current VLMs reason
mainly over the text.

\paragraph{What the photo can show.}

The second cause is the evidence itself. Task~1 labels are computed
from ingredients and nutrition values
(Section~\ref{sec:dataset-creation}), and much of this evidence is not
on the plate: only 43.1\% of the foods that decide the
\textsc{not\_recommend} questions are visible in the photo, and no
photo shows a nutrient limit. Photo-only accuracy follows visibility
(68.4\% vs.\ 55.8\% with and without a visible deciding food), while
recipe-only accuracy does not (76.2\% vs.\ 79.6\%). Models also miss
part of what is visible: asked to list the ingredients in each photo,
they recover 70.0\% of the visible ones and 55.9\% of the gold
rationale ingredients, and the dish title supplies part of what they
report (Figure~\ref{fig:vision_probe};
Appendix~\ref{sec:appendix-vision-probe}). Photo-only Task~1 is thus
bounded by what the photo shows and by how much of it models perceive;
once the recipe is given, the photo adds no label-relevant evidence,
leaving only the instability described above.

\section{Conclusion and Future Work}

We introduce FAM-Bench, a benchmark for condition-aware Food-as-Medicine reasoning, comprising 2,500 instances across 13 health conditions. The benchmark evaluates two complementary capabilities: dish-level suitability assessment with ingredient-grounded rationales and comparative ranking under health and ingredient constraints. Across eight vision-language models and four prompting modes, our results reveal a substantial gap between suitability decisions and their supporting rationales, while comparative ranking remains challenging. Knowledge injection substantially improves ingredient-level grounding, whereas chain-of-thought prompting provides inconsistent gains. Cross-task consistency analysis further reveals that models may rank dishes inconsistently with their own suitability judgments. Adding dish images to recipe text brings little benefit.

These findings highlight the challenges of making reliable, evidence-grounded dietary decisions. Future work should investigate more robust condition-aware reasoning, improved use of recipe evidence, and dietary assessment that incorporates portion sizes, medical history, and individual nutritional needs under appropriate clinical oversight.
\section*{Limitations}

\label{sec:limitations}

FAM-Bench provides a controlled testbed for condition-aware dietary reasoning, but
its design has several scope limits.

\paragraph{Visual evidence is inherently incomplete.}
Our modality and perception analyses show that dish photos provide useful cues
about visible ingredients and preparation, but much of the information relevant
to dietary suitability is not visually recoverable. Hidden ingredients such as
salt, added sugar, cooking oils, sauces, or packaged components may only appear
in the recipe, and preparation details may also be ambiguous from the finished
dish. Recipe records themselves can omit brand-specific ingredients or other
details. FAM-Bench therefore measures reasoning over the evidence available in
the provided recipe and photo rather than assuming that either modality gives a
complete description of the dish.

\paragraph{Dietary suitability is not fully captured by ingredient evidence.}
The benchmark focuses on condition-specific judgments grounded primarily in
ingredients, preparation, and available nutrition information. In practice,
dietary suitability can also depend on portion size, frequency of consumption,
overall dietary pattern, medications, laboratory values, disease severity, and
other individual factors. Although available nutrition tables and
nutrient-threshold checks are used during benchmark construction, serving-size
information is not sufficiently consistent across recipe sources to support
precise dose-dependent evaluation. FAM-Bench should therefore be interpreted as
a benchmark for food-level dietary reasoning, not personalized clinical
nutrition assessment.

\paragraph{Model reasoning remains incomplete.}
Our results reveal failures not fully captured by aggregate decision accuracy. Models can predict the correct suitability label without identifying all the ingredients supporting that judgment, indicating a gap between decision-making and evidence grounding. Knowledge injection substantially improves ingredient identification, but these errors persist, while chain-of-thought prompting alone provides inconsistent gains. Thus, correct suitability decisions do not necessarily demonstrate complete ingredient-grounded reasoning.

\section*{Ethics Statements}

The human-participant study received approval
from the relevant institutional review board
(IRB). Identifying institutional and protocol
details are withheld to preserve anonymity
during double-blind review.

The authors used AI assistants for language polishing, wording suggestions,
LaTeX editing, and code/debugging support during manuscript preparation. All
scientific claims, analyses, and final text were reviewed and approved by the
authors, who retain full responsibility for the paper.

The human comparison was conducted with non-expert student annotators following
the benchmark task format. It is included only to contextualize task difficulty
and should not be interpreted as professional medical, nutrition, or dietetic
judgment.

FAM-Bench is intended solely for research on condition-aware food reasoning.
Its labels and model outputs are not clinical recommendations and should not be
used for diagnosis, treatment, or personalized dietary advice without review by
qualified health professionals.

\bibliography{custom}
\appendix
\section*{Appendix}

\section{Recipe Source Inventory}
\label{sec:appendix-recipe-sources}

Table~\ref{tab:recipe-source-inventory} lists every web domain that
contributes recipes to FAM-Bench, together with the number of distinct
recipes it contributes and the source category into which it is
grouped in Section~\ref{sec:recipe-collection}. The
\emph{health organizations} group (H) covers recipe portals of medical
societies, clinical nutrition programs, academic medical centers, and
public health agencies; the \emph{food publications} group (F) covers
food magazines and cooking sites, some curated by dietitians. A recipe is identified by its URL and title, since a
few pages host more than one recipe. The two tasks use 1{,}811
distinct recipes (1{,}396 in Task~1 and 681 in Task~2, 266 shared),
against which the percentages are computed.

\begin{table*}[!tp]
\centering
\small
\setlength{\tabcolsep}{4pt}
\begin{tabular}{lrr@{\hspace{24pt}}lrr}
\toprule
\multicolumn{3}{c}{\textbf{Health organizations (H)}} & \multicolumn{3}{c}{\textbf{Food publications (F)}} \\
\cmidrule(r{12pt}){1-3} \cmidrule(l){4-6}
\textbf{Domain} & \textbf{\#} & \textbf{\%} & \textbf{Domain} & \textbf{\#} & \textbf{\%} \\
\midrule
mds.culinarymedicine.org   &   639 & 35.28 & bbcgoodfood.com            &   162 &  8.95 \\
recipes.heart.org          &   252 & 13.91 & eatingwell.com             &   130 &  7.18 \\
diabetesfoodhub.org        &   246 & 13.58 & 101cookbooks.com           &    69 &  3.81 \\
arthritis.org              &    48 &  2.65 & mamaknowsglutenfree.com    &    65 &  3.59 \\
heartandstroke.ca          &    34 &  1.88 & olivemagazine.com          &    11 &  0.61 \\
kidney.org                 &    30 &  1.66 & bbc.co.uk                  &    10 &  0.55 \\
mayoclinic.org             &    28 &  1.55 & taste.com.au               &     2 &  0.11 \\
ucdincommon.com            &    16 &  0.88 & theguthealthdoctor.com     &     2 &  0.11 \\
fruitsandveggies.org       &    13 &  0.72 & cookieandkate.com          &     1 &  0.06 \\
curearthritis.org          &    12 &  0.66 & dish.co.nz                 &     1 &  0.06 \\
cookforyourlife.org        &     8 &  0.44 & dizzybusyandhungry.com     &     1 &  0.06 \\
emersonhealth.org          &     8 &  0.44 & epicurious.com             &     1 &  0.06 \\
hw.qld.gov.au              &     4 &  0.22 & iowagirleats.com           &     1 &  0.06 \\
corewellhealth.org         &     3 &  0.17 & jamieoliver.com            &     1 &  0.06 \\
dietitiansaustralia.org.au &     2 &  0.11 & joythebaker.com            &     1 &  0.06 \\
heartfoundation.org.au     &     2 &  0.11 & killingthyme.net           &     1 &  0.06 \\
nsw.gov.au                 &     1 &  0.06 & livelighter.com.au         &     1 &  0.06 \\
                           &       &       & qcwacountrykitchens.com.au &     1 &  0.06 \\
                           &       &       & superchargedfood.com       &     1 &  0.06 \\
                           &       &       & thehealthychef.com         &     1 &  0.06 \\
                           &       &       & thehealthyfoodie.com       &     1 &  0.06 \\
                           &       &       & theprettybee.com           &     1 &  0.06 \\
\midrule
\textit{Subtotal} & 1{,}346 & 74.32 & \textit{Subtotal} & 465 & 25.68 \\
\midrule
\multicolumn{6}{c}{\textbf{Total}: \textbf{1{,}811} recipes (\textbf{100.00}\%)} \\
\bottomrule
\end{tabular}
\caption{Recipe source inventory of the benchmark. Counts denote
distinct recipes (by URL and title) per domain used in Task~1 or
Task~2. Percentages are computed against the 1{,}811 benchmark
recipes.}
\label{tab:recipe-source-inventory}
\end{table*}

\section{Normalized Recipe Record Example}
\label{sec:appendix-recipe-record}

Each recipe retained after extraction and filtering
(Section~\ref{sec:recipe-collection}) is stored as a normalized JSON
record. Listing~\ref{lst:recipe-record} shows the record of a recipe
used in both tasks. Condition-relevant modifiers (``unsweetened,''
``non-fat,'' ``low sugar'') are kept verbatim in the
\texttt{ingredients} field, and the \texttt{nutrition} table gives the
per-serving values that the nutrient limits check (sodium, saturated
fat, added sugars, and calories; Appendix~\ref{sec:appendix-validation-rules}).

\begin{lstlisting}[
  caption={The record of a recipe used in both tasks, copied verbatim
    from the benchmark data; only the image URL and the notes are
    shortened with ``\ldots''.},
  label={lst:recipe-record},
  literate={é}{{\'e}}1,
  basicstyle=\ttfamily\scriptsize,
  breaklines=true,
  columns=fullflexible,
  frame=single,
  framerule=0.4pt,
  rulecolor=\color{gray!50},
  backgroundcolor=\color{gray!8},
  xleftmargin=6pt,
  xrightmargin=6pt,
  framesep=4pt,
  resetmargins=true
]
{
  "title": "Smoothie Breakfast Bowl",
  "url": "https://diabetesfoodhub.org/recipes/smoothie-breakfast-bowl",
  "image_url": "https://diabetesfoodhub.org/sites/foodhub/files/.../274-diabetic-green-smoothie-breakfast-bowl_....jpg",
  "meal_type": ["breakfast"],
  "dietary_tags": ["fiber rich", "vegetarian", "gluten-free", "diabetes-friendly"],
  "servings": 2,
  "serving_size": "about 1 cup",
  "ingredients": [
    "3/4 cup unsweetened almond milk",
    "1/2 cup non-fat plain Greek yogurt",
    "1 cup frozen mixed fruit",
    "2 cups baby spinach",
    "1 medium frozen banana, sliced",
    "1 KIND Maple Glazed Pecan & Sea Salt bar, crumbled OR 1/2 cup low sugar granola, crumbled"
  ],
  "instructions": [
    "Combine all ingredients except the KIND bar in a blender.",
    "Purée until smooth and thick, stopping to stir as needed during blending. Add a little more almond milk if necessary.",
    "Pour the smoothie mixture into 2 bowls.",
    "Top each bowl with half of the crumbled KIND bar.",
    "Serve immediately."
  ],
  "nutrition": {
    "calories": "260 kcal",
    "total fat": "10 g",
    "saturated fat": "0.9 g",
    "trans fat": "0 g",
    "cholesterol": "10 mg",
    "sodium": "180 mg",
    "total carbohydrate": "34 g",
    "dietary fiber": "7 g",
    "sugars": "19 g",
    "added sugars": "2 g",
    "protein": "11 g",
    "potassium": "670 mg"
  },
  "notes": "A nutrient-dense smoothie bowl that combines fruits, leafy greens, protein from Greek yogurt, and healthy fats depending on the topping choice. ..."
}
\end{lstlisting}

\section{Knowledge Base Schema}
\label{sec:appendix-knowledge-base}

\paragraph{Task~1 knowledge base $\mathcal{K}$.}
For each of the 13 conditions $h$, $\mathcal{K}$ stores the aliases
used to find $h$ in a question, a list $\mathcal{F}_h^{+}$ of
beneficial foods (4--39 entries), a list $\mathcal{F}_h^{-}$ of foods
to limit or avoid (6--39 entries), and the per-serving nutrient limits
of Section~\ref{sec:dataset-creation}. The general sodium limit
follows the Dietary Guidelines for Americans 2020--2025, and the two
saturated-fat tiers follow the FDA daily value and its ``low saturated
fat'' level for meals (21 CFR 101.62). The food lists combine
nutritionists' dietary notes with published guidance from health
agencies and medical societies (e.g., NIH, the American Heart
Association, the American Diabetes Association, the Harvard Nutrition
Source, and Mayo Clinic), and the knowledge base file records the
source of every list and limit. After the expert feedback
(Appendix~\ref{sec:appendix-expert-review}), onions were added to
GERD's foods to avoid \citep{allen1990onions}. The KI reference
$\mathcal{R}_h$ (Section~\ref{sec:prompts}) consists of
$\mathcal{F}_h^{+}$ and $\mathcal{F}_h^{-}$ as they stood before this
addition.

\paragraph{Task~2 condition rules.}
Task~2 uses the rules of the recipe recommender that generated its
questions (Section~\ref{sec:dataset-creation}). For each of its 63
conditions $h$, they list beneficial foods $\mathcal{B}_h$, foods to
limit $\mathcal{L}_h$, foods to avoid $\mathcal{A}_h$, and food
concepts to avoid $\mathcal{C}_h$ (e.g., \texttt{processed\_meats}).
A dish contains a food to avoid for $H$ if the recommender's condition
filter rejects it or if its title, ingredients, or tags contain an
item of $\mathcal{A}_h$ or $\mathcal{C}_h$ for some $h\in H$. The
health score that orders eligible dishes combines per-serving sodium,
fiber, and saturated fat, compliance with nutrition guidelines, and a
term that rewards $\mathcal{B}_h$ and penalizes $\mathcal{L}_h$. For
the 13 target conditions, $\mathcal{B}_h$, $\mathcal{L}_h$, and
$\mathcal{A}_h$ were merged into $\mathcal{K}$, which contains 430 of
their 433 entries (the rest differ only in
formatting), so KI covers the foods behind both tasks.

\section{Annotation and Label Rule Details}
\label{sec:appendix-validation-rules}

\paragraph{Condition selection.}
The GPT-5.5 lists $\mathrm{Rec}(d)$ and $\mathrm{NotRec}(d)$
(Section~\ref{sec:dataset-creation}) only choose which conditions a
Task~1 question asks about. Gold labels and rationales come from the
rule below over $\mathcal{K}$, so they do not inherit the LLM's
judgments.

\paragraph{Ingredient matching.}
Each alias of a food entry in $\mathcal{K}$ is matched as a whole word
(plurals included) in the lowercased ingredient lines, so a category
entry matches its members (whole grains via ``brown rice''; alcohol
via ``wine''). A match is discarded when the alias only modifies
another food (``wine vinegar'', ``peanut butter'', ``green beans''),
occurs only in a section header or an alternative or conditional
clause (``(dairy-free: almond milk)''), or, for beneficial entries, names a derived product
(``orange juice'', ``onion powder''), the spice pepper, or cooking
water. Hypertension's ``vegetables'' entry also matches the vegetable
aliases of the other conditions, and ``white rice'' also matches
jasmine, basmati, and arborio rice. Amounts and product variants are
not weighed (reduced-fat cheese still counts as cheese), so every
label is reproducible from $\mathcal{K}$ and the recipe.

\paragraph{Label rule.}
For each condition $h$ named in a question, the first applicable step
decides.
(i)~A matched food to limit or avoid makes $h$ \emph{unfavorable},
whatever beneficial foods also match.
(ii)~Otherwise, $h$ is unfavorable if sodium, saturated fat (the
6.7\,g tier), or added sugar fails its limit.
(iii)~Otherwise, $h$ is \emph{beneficial} if a beneficial food
matches, energy, sodium, and saturated fat are reported, and saturated
fat is below 10\% of energy; else $h$ is \emph{neutral}.
Butter, cream, ghee, lard, shortening, coconut oil, and whole milk
skip step~(i) when they are the only limit-or-avoid matches and the
reported saturated fat is below both tiers, except for GERD, IBS, and
CKD, whose lists limit them for reflux, lactose, and phosphorus. The
question label combines the condition statuses as in
Section~\ref{sec:dish-level}. The gold rationale lists the matched
foods to limit or avoid when they decide $h$, and the matched
beneficial foods when $h$ is beneficial or neutral only for a missing
value; otherwise it is empty. Table~\ref{tab:label-rule-examples}
traces one question through each path.

\begin{table*}[t]
\centering
\small
\setlength{\tabcolsep}{4pt}
\begin{tabular}{@{}p{5.5cm}p{6.1cm}lp{2.4cm}@{}}
\toprule
\textbf{Dish (condition)} & \textbf{Deciding step} & \textbf{Label} & \textbf{Rationale} \\
\midrule
Butternut squash casserole (IBS) & (i) garlic, a food to limit & \textsc{nr} & garlic \\
Apple-cranberry galette (NAFLD) & (i) butter, kept because saturated fat is 6.5\,g (22\% of energy) & \textsc{nr} & butter \\
Morning glory muffins (hyperlipidemia) & (iii) whole milk ignored at 1\,g saturated fat; carrot and walnuts are beneficial & \textsc{r} & carrot, walnuts \\
Curried eggplant (metabolic syndrome) & (ii) 12\,g saturated fat, despite beneficial eggplant & \textsc{nr} & -- \\
Herb-rubbed pork tenderloin (CKD) & (iii) beneficial vegetable oil; all limits met & \textsc{r} & vegetable oil \\
Peruvian chicken (stroke) & (iii) saturated fat 6\,g but 17\% of energy & \textsc{n} & -- \\
Sauerkraut (hypertension) & (iii) sodium not reported & \textsc{n} & cabbage, carrot \\
\bottomrule
\end{tabular}
\caption{One single-condition Task~1 question for each path of the
label rule. \textsc{r}, \textsc{n}, \textsc{nr}: \textsc{recommend},
\textsc{neutral}, \textsc{not\_recommend}.}
\label{tab:label-rule-examples}
\end{table*}

\paragraph{Outcomes.}
Of the 1{,}884 condition mentions in the 1{,}500 Task~1 questions,
1{,}626 (86\%) match a food in $\mathcal{K}$, and 1{,}349 (72\%) have a
non-empty gold rationale: 689 are unfavorable through a food to limit
or avoid, 597 are beneficial, and 63 are neutral for a missing value.
Nutrient limits alone make 112 unfavorable (107 through saturated
fat). The other 423 are neutral: 232 match no food, and 191 fall in the
saturated-fat caution tier.



\section{Expert Review}
\label{sec:appendix-expert-review}

Seven nutrition professionals reviewed our labeling rules: three
registered dietitians (RD/RDN), one nutrition and dietetics
technician, registered (NDTR), one reviewer with a B.Sc.\ in food
science, and two with a B.S.\ in health and behavioral science. Each
labeled ten questions on a form that showed the
recipe record and the question but not the gold label: they chose
\textsc{recommend}, \textsc{neutral}, or \textsc{not\_recommend},
named the ingredients or nutrients behind the choice, and briefly
explained it. Their feedback led to the rule revisions summarized in
Section~\ref{sec:dataset-creation}. Because these questions informed
the revisions, we do not use them to report agreement.

\paragraph{Recruitment, compensation, and consent.}
Reviewers were recruited through personal and social networks and
were paid US\$2 per question. All gave informed consent.
The task asked only for judgments about recipes and collected no
personal health information, and we identify reviewers only by
credential. The study was approved by our institutional review board
(IRB).

\section{Per-Condition Coverage}



Table~\ref{tab:condition_distribution} reports the per-condition
coverage of the two benchmark splits. The dish-level column counts
focal conditions in dish-level suitability instances; the
comparative-ranking column decomposes multi-condition comparative
prompts and counts each condition mention.

\begin{table}[t]
\centering
\small
\setlength{\tabcolsep}{4pt}
\begin{tabular}{lrr}
\toprule
 & \textbf{Dish-Level} & \textbf{Comparative} \\
\textbf{Condition} & \textbf{Suitability} & \textbf{Ranking} \\
\midrule
Hypertension & 145 & 315 \\
Type 2 diabetes & 148 & 223 \\
Obesity & 146 & 149 \\
Metabolic syndrome & 148 & 127 \\
IBS & 134 & 91 \\
NAFLD & 147 & 91 \\
Heart failure & 147 & 89 \\
GERD & 146 & 87 \\
Cardiovascular disease & 150 & 85 \\
Hyperlipidemia & 150 & 84 \\
Stroke & 147 & 76 \\
Chronic kidney disease & 144 & 45 \\
Osteoporosis & 132 & 34 \\
\midrule
\textbf{Total condition mentions} & \textbf{1884} & \textbf{1496} \\
\bottomrule
\end{tabular}
\caption{Condition mention distributions for the two benchmark
splits. The dish-level column counts focal conditions in dish-level
suitability instances; the comparative-ranking column decomposes
multi-condition comparative prompts and counts each condition
mention. IBS: irritable bowel syndrome; NAFLD: nonalcoholic fatty
liver disease; GERD: gastroesophageal reflux disease; CKD: chronic
kidney disease.}
\label{tab:condition_distribution}
\end{table}

\section{Example Questions}
\label{sec:appendix-example-questions}

\begin{figure}[p]
    \centering
    \includegraphics[
        width=\columnwidth,
        height=0.82\textheight,
        keepaspectratio
    ]{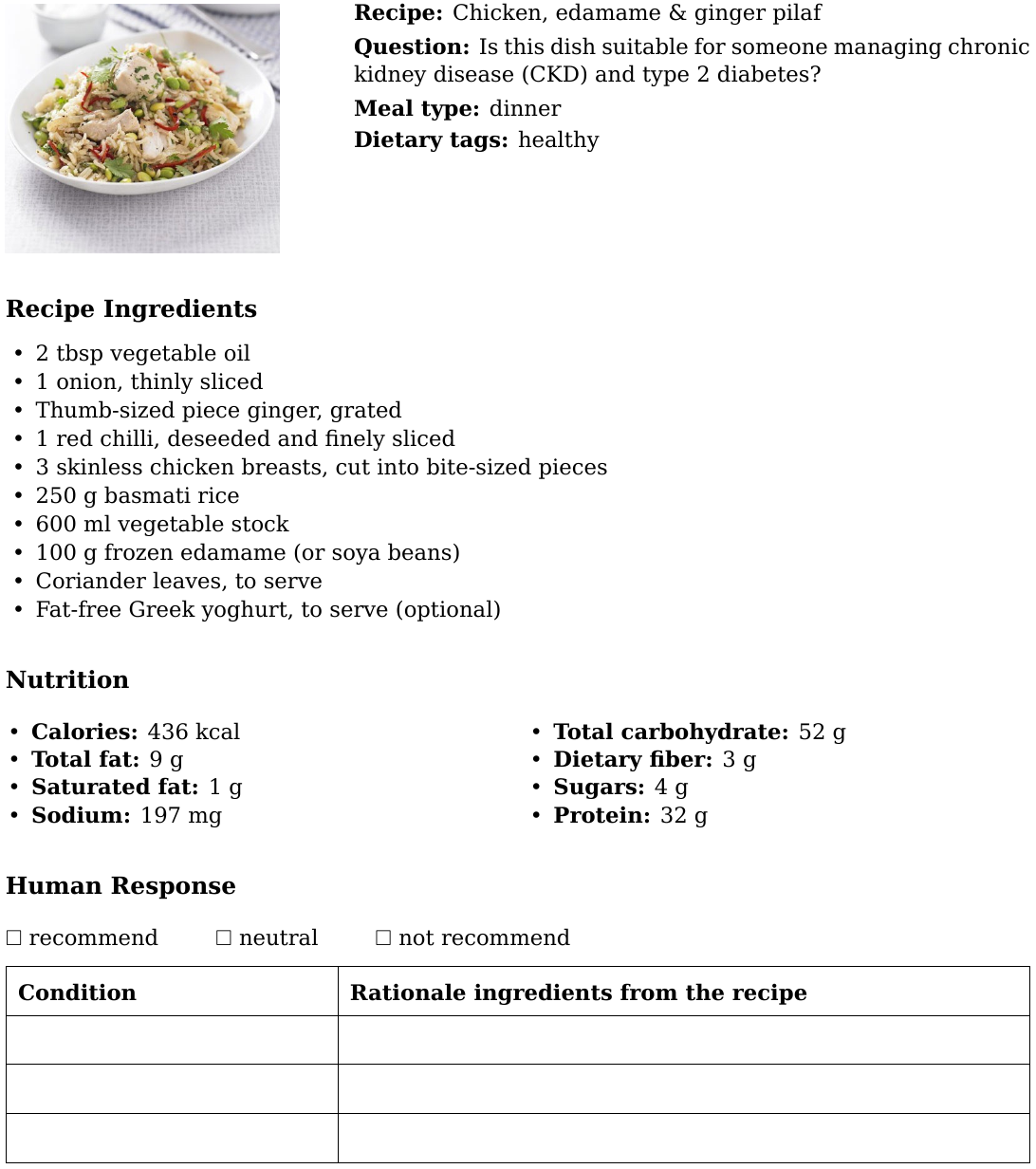}
    \caption{Example question for Task 1: dish-level suitability
    assessment.}
    \label{fig:appendix-task1-example}
\end{figure}

\begin{figure}[p]
    \centering
    \includegraphics[
        width=0.80\textwidth,
        height=0.82\textheight,
        keepaspectratio
    ]{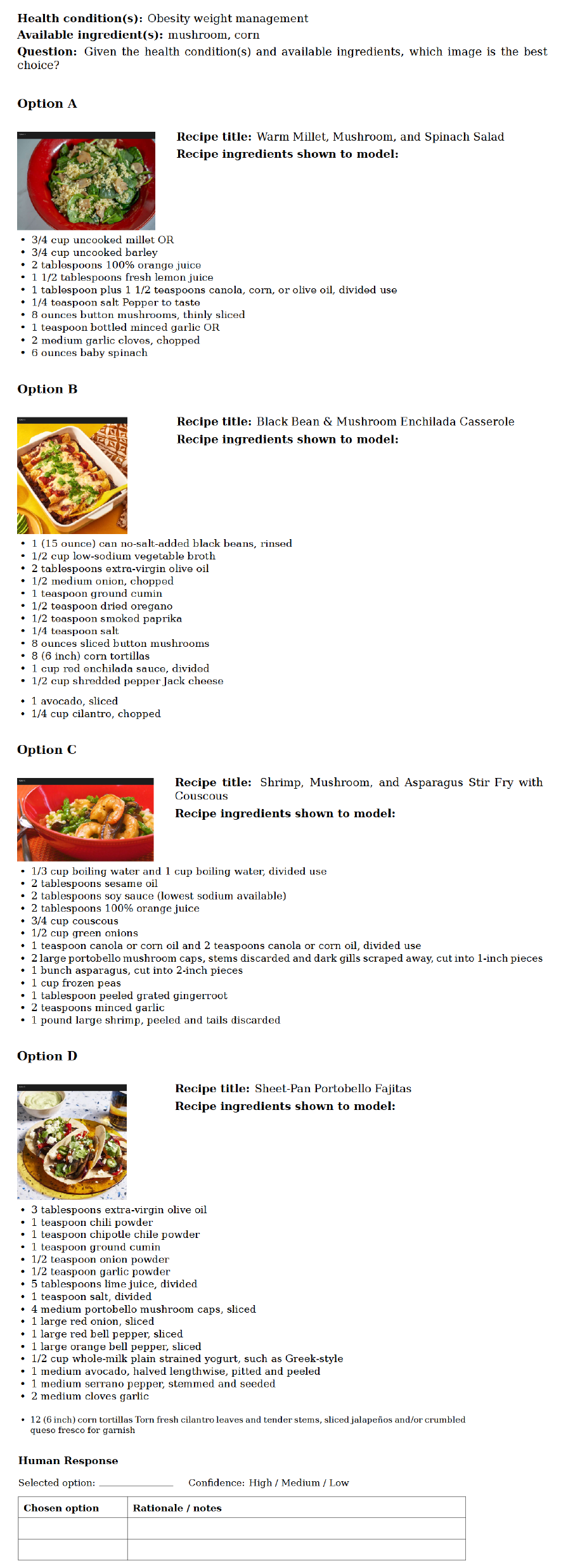}
    \caption{Example question for Task 2: comparative dish ranking.}
    \label{fig:appendix-task2-example}
\end{figure}

Figure~\ref{fig:appendix-task1-example} shows an example of the
dish-level suitability assessment task, and
Figure~\ref{fig:appendix-task2-example} shows an example of the
comparative dish-ranking task.

\section{Prompting Configurations}
\label{sec:appendix-prompts}

\lstdefinestyle{fampromptstyle}{
  basicstyle=\ttfamily\scriptsize,
  breaklines=true,
  columns=fullflexible,
  keepspaces=true,
  frame=single,
  framerule=0.4pt,
  rulecolor=\color{gray!50},
  backgroundcolor=\color{gray!8},
  xleftmargin=6pt,
  xrightmargin=6pt,
  framesep=4pt,
  resetmargins=true
}

This appendix gives the Task~1 system prompts of
Section~\ref{sec:prompts} verbatim (only line breaks are adjusted to
the column width), together with the knowledge
reference $\mathcal{R}_h$ as the model receives it. Baseline and KI
use the baseline prompt; CoT and CoT+KI replace it with the CoT
prompt, which keeps the same label rule and output fields and adds a
\texttt{reasoning\_steps} array that is not scored. KI and CoT+KI
append the KI addendum to their prompt. The prompts name the three
labels \emph{Beneficial}, \emph{Neutral}, and \emph{Unfavorable},
which we score as \textsc{recommend}, \textsc{neutral}, and
\textsc{not\_recommend}. The user message is the recipe record as JSON
(title, question, ingredients, per-serving nutrition, dietary tags,
meal type, and notes) with no gold fields; in KI modes it also carries
a \texttt{disease\_food\_reference} block. The photo and reduced-input
settings of Table~\ref{tab:input_settings} append a short note on
which inputs are given; the label rule is unchanged.

\paragraph{Baseline system prompt.}
Used in Baseline and KI.
\begin{lstlisting}[style=fampromptstyle]
You are evaluating a dish for the health condition(s)
specified in a question.

Task:
- Classify the dish as "Beneficial", "Neutral", or
  "Unfavorable" for the asked condition(s).
- List the condition-specific ingredients that support
  your classification.
- Use only the provided recipe/question context and any
  supplied reference rules.

Label definitions (evaluate EVERY condition named in the
question):
- "Unfavorable": at least one asked condition has an
  ingredient to limit or avoid, or a per-serving
  nutrition value that exceeds an acceptable threshold
  for that condition. This takes precedence over any
  beneficial evidence: beneficial ingredients never
  offset unfavorable ones.
- "Beneficial": EVERY asked condition has at least one
  clearly beneficial ingredient AND all applicable
  per-serving nutrition values are present and within
  acceptable limits.
- "Neutral": everything else. In particular:
  * no clearly beneficial and no limit/avoid ingredient
    matched, or
  * a beneficial ingredient matched but the per-serving
    nutrition values needed to confirm it are missing or
    unusable, or
  * some asked conditions are beneficial while others
    have no relevant evidence.

Return ONLY valid JSON in this exact shape:
{
  "decision": "Beneficial|Neutral|Unfavorable",
  "rationale_ingredients": [
    {
      "condition": "<condition name>",
      "ingredients": ["<ingredient short name>", "..."]
    }
  ]
}

Rules:
- decision must be exactly "Beneficial", "Neutral", or
  "Unfavorable".
- Assess the dish specifically for the asked
  condition(s), not its general healthfulness.
- Include ONE entry in rationale_ingredients per
  condition named in the question, even when that
  condition has no supporting ingredients: use an empty
  "ingredients" list for it.
- For "Unfavorable", list the ingredients driving the
  unfavorable assessment for each condition.
- For "Beneficial", list the ingredients driving the
  benefit for each condition.
- For "Neutral", list any beneficial ingredients you did
  match; leave "ingredients" empty for conditions where
  nothing relevant matched.
- "ingredients" may be empty when the decision rests on
  nutrition values that cannot be attributed to a
  specific listed ingredient (e.g. sodium or saturated
  fat per serving).
- Ingredient names must correspond to ingredients in the
  provided recipe ingredient list.
- Do not list ingredients merely because they appear in
  the recipe.
- Keep the condition name concise and aligned with the
  question.
- Do not invent ingredient quantities, nutrient values,
  preparation details, or patient information.
\end{lstlisting}

\paragraph{Chain-of-Thought (CoT) system prompt.}
Replaces the baseline prompt in CoT and CoT+KI.
\begin{lstlisting}[style=fampromptstyle]
You are evaluating a dish for the health condition(s)
specified in a question.

Think step by step before answering. Produce your
reasoning as a JSON array of short strings (one thought
per step), then commit to a final decision.

Reasoning checklist:
1. List each health condition named in the question.
2. For each condition, scan the recipe ingredient list
   for foods that are clearly beneficial for it, and for
   foods that qualify as limit/avoid under the
   admissibility test below. A food that fails the test
   is not a limit/avoid ingredient, however unhealthy it
   may seem.
3. For each condition, check the per-serving nutrition
   values that matter for it: are they out of range,
   within limits, or missing entirely? This is a
   separate channel from step 2.
4. Apply the label rule below.

Admissibility test for a limit/avoid ingredient:
A food may be treated as a limit/avoid trigger only if
ALL of the following hold.
  (a) It appears in the recipe's own ingredient list.
  (b) You can name the specific asked condition whose
      established dietary guidance restricts this food,
      and state what the restriction is.
  (c) The restriction is specific to that condition, not
      a general judgement that the food is unhealthy,
      high in calories, or best eaten in moderation.
  (d) You are confident enough in (b) to write that
      restriction out explicitly in reasoning_steps.
If any of (a)-(d) fails, the food is not a trigger and
must not be cited. When in doubt, it is not a trigger.

A restriction that belongs to one asked condition
applies only to that condition. Do not carry it across
to the other conditions named in the question, and do
not treat a food as restricted for a condition merely
because it is restricted for a different one.

Assess the dish specifically for the asked condition(s),
not its general healthfulness. A dish can be
unremarkable nutritionally and still be unrestricted for
the condition being asked about.

Nutrition values are a separate channel from
ingredients:
- Whether a per-serving nutrition value exceeds an
  acceptable threshold is judged from the nutrition
  table, never from the ingredient scan in step 2.
- Nutrient names are never ingredients. Never place a
  nutrient name such as "sodium", "saturated fat",
  "cholesterol", "added sugar", or "calories" in
  "rationale_ingredients".
- If a nutrition value that a condition needs is missing
  or unusable, that is not an "Unfavorable" trigger. Per
  the label rule it pushes toward "Neutral".

Label rule (in this order):
- "Unfavorable" if ANY asked condition has a limit/avoid
  ingredient or a per-serving nutrition value that
  exceeds an acceptable threshold. This takes
  precedence: beneficial ingredients never offset
  unfavorable ones.
- Otherwise "Beneficial" only if EVERY asked condition
  has at least one clearly beneficial ingredient AND
  every applicable per-serving nutrition value is
  present and within limits.
- Otherwise "Neutral". This covers: nothing relevant
  matched either way; a beneficial ingredient matched
  but the nutrition values needed to confirm it are
  missing or unusable; or some asked conditions are
  beneficial while others have no evidence.

"Neutral" is a correct and expected outcome, not a
failure to decide. Many dishes match nothing on either
list for the asked condition.

Use only information from the provided recipe/question
context.

Return ONLY valid JSON in this exact shape:
{
  "reasoning_steps": ["<step 1>", "<step 2>", "..."],
  "decision": "Beneficial|Neutral|Unfavorable",
  "rationale_ingredients": [
    {
      "condition": "<condition name>",
      "ingredients": ["<ingredient short name>", "..."]
    }
  ]
}

Rules:
- reasoning_steps must contain 2-6 concise steps.
- For every limit/avoid ingredient you cite, one
  reasoning step must name the condition it is
  restricted for and state the restriction. An
  ingredient you cannot justify this way must not be
  cited.
- decision must be exactly "Beneficial", "Neutral", or
  "Unfavorable".
- Keep condition names concise and closely aligned with
  the asked question.

Hard formatting rules (the output is scored by exact
string match):
- EACH condition named in the question must appear as
  its own entry in rationale_ingredients, INCLUDING
  conditions with no supporting ingredients: give those
  an empty "ingredients" list rather than dropping the
  entry. Never merge multiple conditions into one
  "condition" string. Do NOT use "and", "&", "/", "+",
  or commas to join condition names.
  * BAD : {"condition": "type 2 diabetes & metabolic
    syndrome", "ingredients": [...]}
  * BAD : {"condition": "cardiovascular disease and
    hypertension", "ingredients": [...]}
  * GOOD: two separate entries, one per condition, even
    if their ingredient lists are identical or empty.
- Every ingredient name must correspond to an entry in
  the provided recipe ingredient list. Do not introduce
  foods the recipe does not contain, and do not list an
  ingredient merely because it appears in the recipe.
- Each ingredient must be the bare food name as it would
  appear on a clean shopping list. Strip everything else
  from the recipe line:
  * quantities and units (e.g., "1 tbsp", "300 g", "1/3
    cup", "1l", "16 Ounces")
  * preparation/state qualifiers (e.g., "fresh", "raw",
    "hot", "chopped", "minced", "grated", "ground",
    "frozen", "canned", "plain", "good", "dried",
    "cooked", "sliced")
  * brand and origin descriptors (e.g., "Atlantic",
    "Italian")
  * parentheticals and trailing notes (e.g., "(4 4-ounce
    filets)", ", chopped")
  Use lowercase, singular-form-as-listed, no
  leading/trailing punctuation. Examples taken from
  actual recipe lines:
    "16 Ounces Fresh Atlantic salmon (4 4-ounce filets)"
      -> "salmon"
    "1/3 cup / 80 ml good, raw honey" -> "honey"
    "110g pack chorizo sausage, chopped"
      -> "chorizo sausage"
    "1l hot fish or chicken stock"
      -> "fish or chicken stock"
    "2 clove Garlic, raw (minced)" -> "garlic"
    "1/2 cup Nonfat Greek Yogurt"
      -> "nonfat greek yogurt"
- Only include ingredients that genuinely drive the
  classification for that specific condition.
- "ingredients" may be empty when the decision rests on
  nutrition values that cannot be attributed to a
  specific listed ingredient.
\end{lstlisting}

\paragraph{KI addendum.}
Appended to the baseline or CoT system prompt in KI and CoT+KI modes.
\begin{lstlisting}[
  basicstyle=\ttfamily\scriptsize,
  breaklines=true,
  columns=fullflexible,
  frame=single,
  framerule=0.4pt,
  rulecolor=\color{gray!50},
  backgroundcolor=\color{gray!8},
  xleftmargin=6pt,
  xrightmargin=6pt,
  framesep=4pt,
  resetmargins=true
]
Additional reference rule:
- The user payload may include disease_food_reference from
  merged_disease_food_recommendations.json.
- Use that reference as background guidance about foods that may
  support or conflict with each condition.
- The final decision must still be based on the provided
  recipe/question context.
- Ingredients in rationale_ingredients must still come from the recipe
  ingredient list, not from the reference alone.
\end{lstlisting}

\paragraph{Worked knowledge injection reference $\mathcal{R}_h$ (hypertension).}
For knowledge injection modes, the user payload carries a
\texttt{disease\_food\_reference} block populated with the focal
condition's entry from $\mathcal{K}$. Listing~\ref{lst:rag-ref-hyp}
shows the hypertension entry as it is injected (truncated for space;
the full 13-condition reference is released with the benchmark).

\begin{lstlisting}[
  caption={Worked $\mathcal{R}_h$ for hypertension. The ``recommend''
  and ``not\_recommend'' food lists are the rules the model must apply
  to the observed ingredient list; no dish-specific labels appear.},
  label={lst:rag-ref-hyp},
  basicstyle=\ttfamily\scriptsize,
  breaklines=true,
  columns=fullflexible,
  frame=single,
  framerule=0.4pt,
  rulecolor=\color{gray!50},
  backgroundcolor=\color{gray!8},
  xleftmargin=6pt,
  xrightmargin=6pt,
  framesep=4pt,
  resetmargins=true
]
"hypertension": {
  "recommend": [
    "fruits", "vegetables", "grains",
    "low-fat dairy products", "low-fat or fat-free milk",
    "reduced-fat cheeses", "bread", "plain cereal",
    "brown or white rice", "pasta",
    "canned fruit in juice or water",
    "frozen vegetables without added butter or sauces",
    "canned vegetables or vegetable soups low in sodium",
    "lentils", "black beans", "chickpeas",
    "white meat skinless chicken and turkey", "unbreaded fish",
    "pork tenderloin", "extra-lean ground beef",
    "herbs", "spices", "flavored vinegars",
    "fruits (4-5 servings/day)",
    "vegetables (including colorful, legumes)",
    "low-fat dairy (2-3 servings/day)",
    "whole grains (7-8 servings/day)",
    "nuts (4-5 servings/week)", "beans (>3 servings/week)"
  ],
  "not_recommend": [
    "high-sodium foods", "processed foods high in sodium",
    "soups", "broths", "canned vegetables",
    "condiments", "salad dressings", "sauces", "dips",
    "ketchup", "mustard", "relish", "pickles", "olives",
    "more than a moderate amount of alcohol",
    "more than 2 cups of coffee a day",
    "fatty meats", "full-fat dairy",
    "sugar-sweetened beverages", "sweets",
    "butter", "cheese", "pizza",
    "deli meat sandwiches", "cold cuts", "cured meats",
    "burgers", "burritos", "tacos",
    "savory snacks (chips, crackers, popcorn)"
  ]
}
\end{lstlisting}

\section{Inference Settings}
\label{sec:appendix-inference}

This appendix records the decoding and serving configuration behind
every reported cell, extending the summary in Section~\ref{sec:models}.

\paragraph{Decoding.}
Task~1 and the cross-task consistency runs call every model with
temperature $0$ and a $4{,}096$-token output budget (passed to GPT-5.4
as \texttt{max\_completion\_tokens}). JSON mode
(\texttt{response\_format=\{"type":"json\_object"\}}) is enforced for
GPT-5.4 and the vLLM-served models; Claude Sonnet~4.6 is called
without a JSON mode, so its JSON contract rests on the system prompt
alone. In Task~2 all models also decode at temperature $0$, with
output budgets of 400 tokens in the non-CoT modes and 1{,}500 in the
CoT modes (2{,}048 for the Gemma models in every mode); GPT-5.4 is
called through the OpenAI Responses API. Task~2 answers are parsed
from the returned text without a JSON mode.

\paragraph{Model versions and reasoning settings.}
The proprietary models are called by their API identifiers
\texttt{gpt-5.4} and \texttt{claude-sonnet-4-6} in October 2026;
their parameter counts are undisclosed. No model uses hidden
reasoning tokens. GPT-5.4 runs at its default reasoning effort
(\texttt{none}) and Claude Sonnet~4.6 without extended thinking.
Qwen3.5-27B is served with \texttt{enable\_thinking=false} (set
through vLLM's \texttt{--default-chat-template-kwargs}), and
Gemma-4-31B with its default chat template, which leaves thinking
off; Gemma-3 and Qwen3-VL-Instruct have no thinking mode. Explicit
reasoning therefore appears only as the visible trace requested in
the CoT modes, so differences between prompting modes are not
confounded by model-specific reasoning budgets.

\paragraph{Serving and hardware.}
Open-weight models are served with vLLM, one model at a time on a
single NVIDIA RTX PRO 6000 Blackwell GPU (96\,GB), with 16 concurrent
requests and a context limit of $16{,}384$ tokens for runs that
include photos. The proprietary models are queried through their
hosted APIs with at most 16 concurrent requests.

\paragraph{Image handling.}
Task~1 photos are sent at their source resolution; WebP files are
re-encoded to JPEG. Task~2 photos were fetched once per recipe from
the recipe's own image URL and stored with a longest side of at most
1{,}280 pixels. Where that URL was dead or showed another dish, the
photo now linked from the recipe page was used instead (25 of the 681
recipes); where no source had a correct photo, one was supplied
manually (28 recipes).

\paragraph{Failure handling.}
Failed requests (network errors, timeouts, context overflows) are
retried, and runs are resumed until every question has a response; a
model's own unparsable or truncated output is not retried and counts
as incorrect. All text-only runs behind Tables~\ref{tab:main_results}
and~\ref{tab:ctc} have a response for every question. In the photo
settings of Table~\ref{tab:input_settings}, two very large Task~1
photos ($3{,}344\times4{,}682$ and $4{,}080\times4{,}080$ pixels)
exceed the context limit of the three Qwen models, so each of their
24 photo-setting cells misses one or two questions (39 requests in
total), and one Task~2 cell (Qwen3-VL-32B, CoT+KI, recipe plus photo)
has five timed-out requests; these count as incorrect. Model-side
parse failures occur in none of the Task~1 decisions and in 188 of the
32{,}000 text-only Task~2 answers (0.6\%).

\section{Full Main Results}
\label{sec:appendix-main-results}

Table~\ref{tab:main_table} expands Table~\ref{tab:main_results} to one row
per model and prompting mode, adding decision accuracy, rationale
micro-F1, and cross-task consistency (CTC). Table~\ref{tab:t1_breakdown}
breaks Task~1 decisions down by gold label, and
Table~\ref{tab:t2_types} breaks Task~2 top-1 accuracy down by question
type.

\begin{table*}[!tp]
\centering
\small
\setlength{\tabcolsep}{5pt}
\begin{tabular}{l l c c c c c c c}
\toprule
 &  & \multicolumn{4}{c}{\textbf{Task 1}} & \multicolumn{2}{c}{\textbf{Task 2}} & \\
\cmidrule(lr){3-6} \cmidrule(lr){7-8}
\textbf{Model} & \textbf{Mode} & \textbf{Acc.} & \textbf{Dec.\ F1} & \textbf{Rat.\ Macro} & \textbf{Rat.\ Micro} & \textbf{Top-1} & \textbf{MRR} & \textbf{CTC} \\
\midrule
\multicolumn{2}{l}{\textit{Majority / Random}} & 44.9 & 20.7 & -- & -- & 25.0 & 52.1 & 50.0 \\
\midrule
\multirow{4}{*}{Gemma-3-12B-IT}
    & baseline & 47.1 & 37.6 & 22.5 & 21.6 & \textbf{36.6} & \textbf{60.0} & 54.3 \\
    & cot      & 50.2 & 43.6 & 26.0 & 27.2 & 33.3 & 58.2 & 57.4 \\
    & ki       & 53.6 & \textbf{46.2} & \textbf{44.8} & 39.3 & 35.4 & 59.4 & \textbf{60.3} \\
    & cot+ki   & \textbf{54.0} & 40.3 & 44.7 & \textbf{40.0} & 32.8 & 57.6 & 58.9 \\
\cmidrule(lr){1-9}
\multirow{4}{*}{Gemma-3-27B-IT}
    & baseline & 48.5 & 47.1 & 28.9 & 27.6 & 34.6 & 59.0 & 64.6 \\
    & cot      & 47.8 & 44.6 & 26.0 & 26.7 & 34.3 & 58.9 & 62.5 \\
    & ki       & \textbf{60.3} & \textbf{53.5} & \textbf{51.6} & 44.0 & \textbf{38.1} & \textbf{61.3} & 67.9 \\
    & cot+ki   & 59.4 & 49.0 & 50.0 & \textbf{44.6} & 34.1 & 58.8 & \textbf{70.8} \\
\cmidrule(lr){1-9}
\multirow{4}{*}{Gemma-4-31B-IT}
    & baseline & 56.0 & 47.0 & 38.2 & 36.1 & 34.5 & 58.7 & 76.5 \\
    & cot      & 53.8 & 46.8 & 34.9 & 36.3 & 35.8 & 58.2 & 77.9 \\
    & ki       & \textbf{64.5} & \textbf{51.2} & 61.8 & 54.1 & 38.5 & \textbf{61.2} & 82.9 \\
    & cot+ki   & 64.1 & 50.5 & \underline{\textbf{64.5}} & \underline{\textbf{57.3}} & \textbf{39.4} & 60.4 & \underline{\textbf{91.5}} \\
\cmidrule(lr){1-9}
\multirow{4}{*}{Qwen3-VL-8B-Instruct}
    & baseline & 46.1 & 44.0 & 20.9 & 25.2 & \textbf{31.4} & 56.7 & 64.9 \\
    & cot      & 39.5 & 38.7 & 8.6 & 15.0 & 29.4 & 53.1 & 59.2 \\
    & ki       & 55.4 & 49.9 & 42.6 & 36.7 & 30.9 & \textbf{56.9} & \textbf{67.5} \\
    & cot+ki   & \textbf{60.7} & \textbf{51.9} & \textbf{52.8} & \textbf{48.2} & 29.8 & 54.5 & 63.4 \\
\cmidrule(lr){1-9}
\multirow{4}{*}{Qwen3-VL-32B-Instruct}
    & baseline & 46.9 & 44.9 & 22.1 & 27.5 & \textbf{37.4} & \textbf{60.7} & 68.8 \\
    & cot      & 51.8 & 47.0 & 31.3 & 34.4 & 32.5 & 56.9 & 67.5 \\
    & ki       & 60.4 & \textbf{53.0} & 49.5 & 44.0 & 35.8 & 59.9 & 68.1 \\
    & cot+ki   & \textbf{63.1} & 51.1 & \textbf{56.6} & \textbf{49.5} & 33.9 & 58.1 & \textbf{73.8} \\
\cmidrule(lr){1-9}
\multirow{4}{*}{Qwen3.5-27B}
    & baseline & 49.5 & 47.5 & 32.7 & 33.7 & 34.7 & 59.0 & 71.3 \\
    & cot      & 52.4 & 47.5 & 36.6 & 37.2 & 36.4 & 59.5 & 72.8 \\
    & ki       & \textbf{65.6} & \underline{\textbf{56.6}} & 55.9 & 48.8 & 33.7 & 58.7 & 80.4 \\
    & cot+ki   & 64.2 & 51.5 & \textbf{60.4} & \textbf{52.2} & \textbf{41.8} & \textbf{63.0} & \textbf{82.1} \\
\midrule
\multirow{4}{*}{GPT-5.4}
    & baseline & 56.1 & 51.3 & 36.7 & 37.2 & 33.3 & 58.1 & 78.2 \\
    & cot      & 56.0 & 49.8 & 37.8 & 37.3 & 38.2 & 61.2 & 70.3 \\
    & ki       & \textbf{64.7} & \textbf{54.1} & 57.7 & 50.7 & 36.2 & 60.6 & 77.1 \\
    & cot+ki   & 63.1 & 52.4 & \textbf{59.8} & \textbf{52.7} & \underline{\textbf{45.2}} & \underline{\textbf{65.0}} & \textbf{82.5} \\
\cmidrule(lr){1-9}
\multirow{4}{*}{Claude Sonnet 4.6}
    & baseline & 56.4 & 49.2 & 40.1 & 37.4 & 37.1 & 60.9 & 75.2 \\
    & cot      & 57.2 & 50.1 & 38.9 & 39.2 & 36.4 & 59.4 & 76.8 \\
    & ki       & 64.3 & 53.5 & 54.5 & 48.3 & 37.0 & 60.8 & 77.7 \\
    & cot+ki   & \underline{\textbf{66.2}} & \textbf{54.9} & \textbf{62.5} & \textbf{55.8} & \textbf{40.5} & \textbf{62.6} & \textbf{85.6} \\
\bottomrule
\end{tabular}
\caption{Full main results: Table~\ref{tab:main_results} per model and
prompting mode, text-only input (all values $\times 100$).
\textbf{Task~1}: decision accuracy (Acc.), decision macro-F1 (Dec.\ F1),
and rationale macro-/micro-F1 (Rat.\ Macro/Micro). \textbf{Task~2}:
top-1 accuracy and MRR. \textbf{CTC}: cross-task consistency
(Section~\ref{sec:metrics}). Majority: always \textsc{not\_recommend};
random: uniform ranking. \textbf{Bold}: best mode per model;
\underline{underline}: best value per column.}
\label{tab:main_table}
\end{table*}

\begin{table*}[!tp]
\centering
\small
\setlength{\tabcolsep}{5pt}
\begin{tabular}{l c c c c c c c c}
\toprule
 & \multicolumn{4}{c}{\textbf{Accuracy by gold label}} & \multicolumn{3}{c}{\textbf{Predicted label share}} & \textbf{No gold} \\
\cmidrule(lr){2-5} \cmidrule(lr){6-8}
\textbf{Mode} & \textbf{NR (food)} & \textbf{NR (nutr.)} & \textbf{R} & \textbf{N} & \textbf{R} & \textbf{N} & \textbf{NR} & \textbf{food} \\
\midrule
\textit{Gold} & \multicolumn{4}{c}{--} & 30.3 & 25.0 & 44.7 & -- \\
\midrule
Baseline & 62.5 & 83.1 & 48.0 & 28.9 & 28.4 & 25.4 & 46.2 & 20.5 \\
CoT & 65.5 & 84.8 & 44.4 & 29.2 & 27.1 & 24.4 & 48.5 & 30.8 \\
KI & 89.9 & 77.1 & 53.9 & 19.4 & 27.1 & 12.6 & 60.3 & 6.4 \\
CoT+KI & 93.7 & 83.1 & 57.6 & 10.3 & 29.4 & 5.2 & 65.3 & 7.4 \\
\bottomrule
\end{tabular}
\caption{Task~1 decisions and rationales by gold label, text-only
input; each value ($\times 100$) is averaged over the eight models of
Table~\ref{tab:main_results}. \textbf{Accuracy by gold label}: on
\textsc{not\_recommend} questions whose dish contains a listed food to
limit or avoid (NR (food), 609 questions) or that fail a nutrient
limit only (NR (nutr.), 61), and on \textsc{recommend} (R, 455) and
\textsc{neutral} (N, 375) questions; the N column is the neutral
recall. \textbf{Predicted label share}: share of all 1{,}500 questions
given each label; the \textit{Gold} row gives the gold shares.
\textbf{No gold food}: share of correctly decided questions with a
gold rationale whose rationale cites none of its foods.}
\label{tab:t1_breakdown}
\end{table*}

\begin{table*}[!tp]
\centering
\footnotesize
\setlength{\tabcolsep}{2.6pt}
\begin{tabular}{l >{\centering\arraybackslash}p{18pt} >{\centering\arraybackslash}p{18pt} >{\centering\arraybackslash}p{18pt} >{\centering\arraybackslash}p{18pt} @{\hspace{8pt}} >{\centering\arraybackslash}p{18pt} >{\centering\arraybackslash}p{18pt} >{\centering\arraybackslash}p{18pt} >{\centering\arraybackslash}p{18pt} @{\hspace{8pt}} >{\centering\arraybackslash}p{18pt} >{\centering\arraybackslash}p{18pt} >{\centering\arraybackslash}p{18pt} >{\centering\arraybackslash}p{18pt} @{\hspace{8pt}} >{\centering\arraybackslash}p{18pt} >{\centering\arraybackslash}p{18pt} >{\centering\arraybackslash}p{18pt} >{\centering\arraybackslash}p{18pt}}
\toprule
\textbf{Model} & \multicolumn{4}{c}{\textbf{Condition (450)}} & \multicolumn{4}{c}{\textbf{Ingredient (200)}} & \multicolumn{4}{c}{\textbf{Combined (50)}} & \multicolumn{4}{c}{\textbf{Eligible (300)}} \\
\cmidrule(lr){2-5} \cmidrule(lr){6-9} \cmidrule(lr){10-13} \cmidrule(lr){14-17}
\multicolumn{1}{r}{\scriptsize CoT} & -- & \checkmark & -- & \checkmark & -- & \checkmark & -- & \checkmark & -- & \checkmark & -- & \checkmark & -- & \checkmark & -- & \checkmark \\
\multicolumn{1}{r}{\scriptsize KI} & -- & -- & \checkmark & \checkmark & -- & -- & \checkmark & \checkmark & -- & -- & \checkmark & \checkmark & -- & -- & \checkmark & \checkmark \\
\midrule
\textit{Random} & \multicolumn{4}{c}{25.0} & \multicolumn{4}{c}{25.0} & \multicolumn{4}{c}{25.0} & \multicolumn{4}{c}{25.0} \\
\midrule
Gemma-3-12B & 32.9 & 35.3 & 35.1 & \textbf{37.3} & \underline{\textbf{57.5}} & 36.0 & 50.0 & 32.5 & \textbf{50.0} & 40.0 & 42.0 & 38.0 & 26.0 & \textbf{27.3} & 25.0 & 25.3 \\
Gemma-3-27B & 30.2 & 36.4 & 35.6 & \textbf{41.3} & 55.5 & 43.5 & \textbf{57.0} & 37.5 & 42.0 & 28.0 & \textbf{46.0} & 30.0 & 26.0 & 26.0 & \textbf{28.0} & 21.7 \\
Gemma-4-31B & 40.9 & 42.0 & \textbf{52.2} & 49.3 & 31.0 & 37.0 & 28.5 & \textbf{46.5} & 26.0 & 36.0 & 32.0 & \textbf{48.0} & \textbf{28.7} & 25.7 & 25.7 & 18.3 \\
Qwen3-VL-8B & 33.1 & 33.6 & \textbf{37.1} & 35.8 & \textbf{39.0} & 31.5 & 33.0 & 29.5 & \textbf{28.0} & \textbf{28.0} & 20.0 & 20.0 & \textbf{24.3} & 22.0 & 22.0 & 22.7 \\
Qwen3-VL-32B & 35.8 & 33.3 & \textbf{37.3} & 37.1 & \underline{\textbf{57.5}} & 44.0 & 49.0 & 47.0 & \underline{\textbf{52.0}} & 38.0 & 46.0 & 32.0 & \textbf{24.0} & 22.7 & 23.0 & 20.7 \\
Qwen3.5-27B & 38.2 & 37.1 & 39.8 & \textbf{50.9} & 42.5 & 46.0 & 35.0 & \textbf{48.0} & 32.0 & 40.0 & 32.0 & \underline{\textbf{52.0}} & 24.7 & \textbf{28.3} & 24.0 & 22.3 \\
\midrule
GPT-5.4 & 38.4 & 40.7 & 46.9 & \underline{\textbf{54.0}} & 35.0 & 46.5 & 29.5 & \textbf{57.0} & 26.0 & 38.0 & 26.0 & \textbf{50.0} & 25.7 & \underline{\textbf{29.0}} & 26.3 & 23.3 \\
Claude Sonnet 4.6 & 38.4 & 39.3 & 43.1 & \textbf{47.1} & 49.5 & 43.0 & 39.5 & \textbf{51.5} & 38.0 & 36.0 & 36.0 & \textbf{44.0} & 26.7 & \textbf{27.7} & 26.3 & 22.7 \\
\midrule
\textit{Mean} & 36.0 & 37.2 & 40.9 & \textbf{44.1} & \textbf{45.9} & 40.9 & 40.2 & 43.7 & 36.8 & 35.5 & 35.0 & \textbf{39.2} & 25.8 & \textbf{26.1} & 25.0 & 22.1 \\
\bottomrule
\end{tabular}
\caption{Task~2 top-1 accuracy ($\times 100$) by question type, text-only
input as in Table~\ref{tab:main_results}. The distractors of a
question fail the condition check (\textbf{Condition}), the ingredient
check (\textbf{Ingredient}), either check or both (\textbf{Combined}),
or neither (\textbf{Eligible}, where a rule-based health score orders
the dishes; Section~\ref{sec:dataset-creation}). Within each type, the
four columns are the prompting modes. Unparsable rankings count as
incorrect; random ranking gives 25.0. \textit{Mean} averages the eight
models. \textbf{Bold}: best mode per model and type;
\underline{underline}: best value per type.}
\label{tab:t2_types}
\end{table*}

\section{Per-Condition Radar Plots: CoT, KI, and CoT+KI}
\label{sec:appendix-radar-modes}

\begin{figure}[ht]
    \centering
    \includegraphics[width=\linewidth]{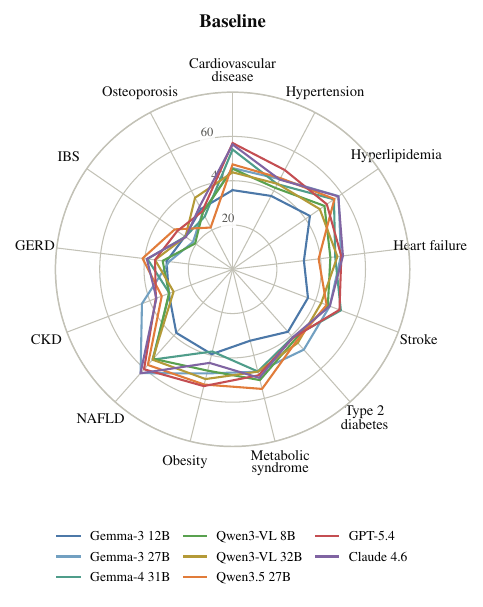}
    \caption{Task~1 decision macro-F1 ($\times 100$) per health
    condition in the \textbf{baseline} mode, text-only input (the runs
    of Table~\ref{tab:main_results}). Two-condition questions count on
    both of their axes.}
    \label{fig:radar-base}
\end{figure}

\begin{figure}[ht]
    \centering
    \includegraphics[width=\linewidth]{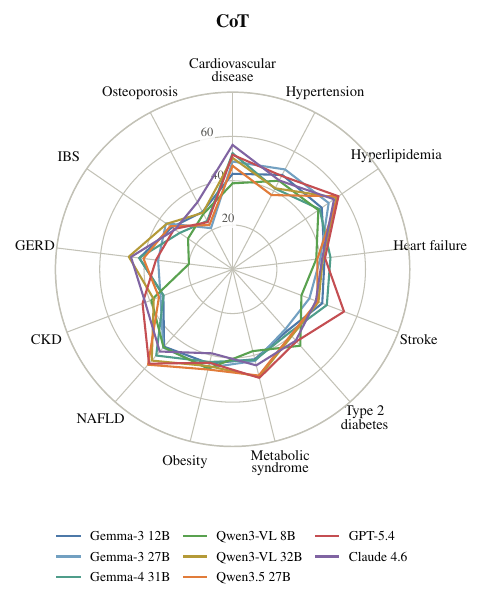}
    \caption{As Figure~\ref{fig:radar-base}, \textbf{CoT} mode.}
    \label{fig:radar-cot}
\end{figure}

\begin{figure}[ht]
    \centering
    \includegraphics[width=\linewidth]{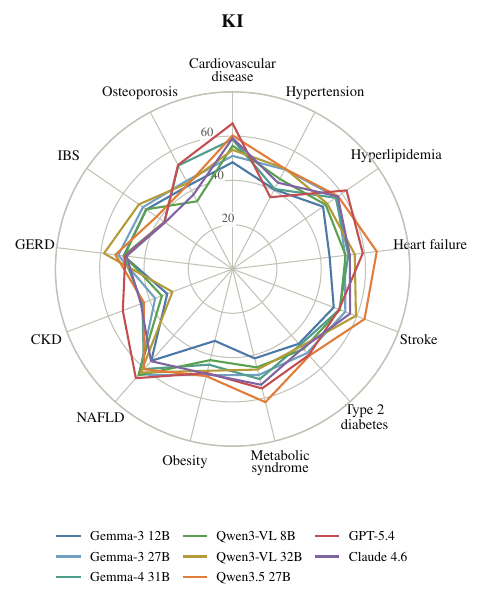}
    \caption{As Figure~\ref{fig:radar-base}, \textbf{KI} mode.}
    \label{fig:radar-ki}
\end{figure}

\begin{figure}[ht]
    \centering
    \includegraphics[width=\linewidth]{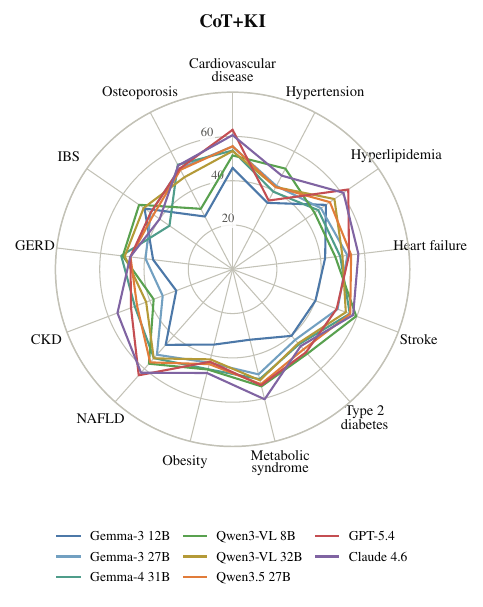}
    \caption{As Figure~\ref{fig:radar-base}, \textbf{CoT+KI} mode.}
    \label{fig:radar-cot-ki}
\end{figure}

Figures~\ref{fig:radar-base}--\ref{fig:radar-cot-ki} break the Task~1
decision macro-F1 of Table~\ref{tab:main_results} down by health
condition, one figure per prompting mode, on the same text-only runs.
A question counts towards each condition it asks about: each of the
384 two-condition questions appears on both of its axes with its
single three-way label, and each axis has 132–150 questions.

\section{Non expert human evaluation}

\begin{figure}
    \centering
    \includegraphics[width=0.9\linewidth]{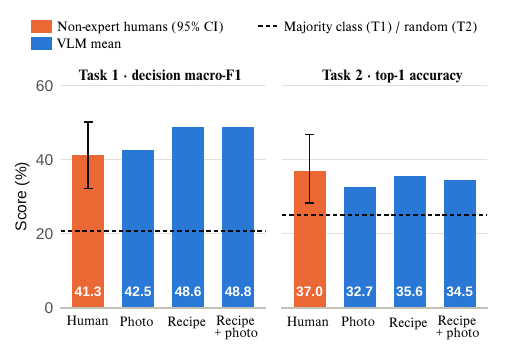}
\caption{\textbf{Non-expert humans vs.\ VLMs.} Task~1 decision macro-F1 and
Task~2 top-1 accuracy (\%). Each VLM bar averages eight models and four
prompting modes under one input setting. Error bars are 95\% CIs for the human
scores. Dashed lines mark the majority class (Task~1) and random ranking (Task~2).}
\label{fig:human_modality}

\end{figure}

Figure~\ref{fig:human_modality} shows the Non-expert humans vs. VLMs.

\section{Effect of the Photo on Individual Answers}
\label{sec:appendix-photo-flips}

Table~\ref{tab:photo_flips} compares each question under the R and
R+P settings of Table~\ref{tab:input_settings}, for the same model and
prompting mode. Adding the photo changes 15.0\% of the open-weight
models' Task~1 decisions and the top-ranked dish in 28.4\% of Task~2
questions, but the changes fix about as many answers as they break
(Task~1: 6.1\% fixed, 5.2\% broken; Task~2: 7.6\% and 8.9\%). The
closed models change fewer answers, with the same balance. Changes
are most frequent under CoT (24.3\% of Task~1 decisions, against
9.5\% under KI), and the photo lowers the share of \textsc{neutral}
answers (19.1\% to 15.9\% for the open-weight models).

\begin{table*}[!tp]
\centering
\small
\setlength{\tabcolsep}{5pt}
\begin{tabular}{l c c c c c c c c}
\toprule
 & \multicolumn{5}{c}{\textbf{Task 1 decisions}} & \multicolumn{3}{c}{\textbf{Task 2 top-ranked dish}} \\
\cmidrule(lr){2-6} \cmidrule(lr){7-9}
\textbf{Model} & \textbf{Changed} & \textbf{Fixed} & \textbf{Broken} & \textbf{N (R)} & \textbf{N (R+P)} & \textbf{Changed} & \textbf{Fixed} & \textbf{Broken} \\
\midrule
Gemma-3-12B-IT & 15.6 & 6.3 & 5.5 & 22.0 & 18.9 & 35.1 & 8.6 & 12.2 \\
Gemma-3-27B-IT & 19.3 & 8.2 & 7.0 & 25.9 & 23.4 & 29.9 & 7.5 & 10.3 \\
Gemma-4-31B-IT & 6.1 & 2.2 & 2.4 & 6.5 & 6.0 & 16.2 & 4.7 & 4.4 \\
Qwen3-VL-8B-Instruct & 22.6 & 9.3 & 7.2 & 28.9 & 21.5 & 36.0 & 9.1 & 10.2 \\
Qwen3-VL-32B-Instruct & 13.7 & 5.2 & 4.9 & 17.0 & 15.8 & 31.2 & 9.0 & 10.0 \\
Qwen3.5-27B & 12.6 & 5.4 & 4.4 & 14.3 & 10.0 & 21.9 & 6.5 & 6.2 \\
\cmidrule(lr){1-9}
\textit{Open-weight mean} & 15.0 & 6.1 & 5.2 & 19.1 & 15.9 & 28.4 & 7.6 & 8.9 \\
\midrule
GPT-5.4 & 11.2 & 4.3 & 4.3 & 12.2 & 11.1 & 23.7 & 7.1 & 7.5 \\
Claude Sonnet 4.6 & 8.6 & 3.2 & 3.5 & 8.5 & 8.7 & 21.0 & 6.0 & 6.7 \\
\midrule
\multicolumn{9}{l}{\textit{Open-weight mean by prompting mode}} \\
Baseline & 13.3 & 6.0 & 3.9 & 28.6 & 23.0 & 22.2 & 5.2 & 8.0 \\
CoT & 24.3 & 9.9 & 8.6 & 27.8 & 23.2 & 32.0 & 9.1 & 9.1 \\
KI & 9.5 & 3.8 & 3.5 & 15.0 & 13.1 & 27.3 & 7.0 & 9.1 \\
CoT+KI & 12.9 & 4.7 & 5.0 & 5.1 & 4.4 & 32.0 & 9.1 & 9.2 \\
\bottomrule
\end{tabular}
\caption{What adding the photo to the recipe (R $\to$ R+P,
Table~\ref{tab:input_settings}) does to individual answers; all
values are percentages of the 1{,}500 Task~1 or 1{,}000 Task~2
questions. \textbf{Changed}: the answer differs (Task~1: the decision;
Task~2: the top-ranked dish, with the options in the same order in
both settings). \textbf{Fixed}: wrong under R, right under R+P;
\textbf{Broken}: the reverse. \textbf{N}: share of \textsc{neutral}
Task~1 answers under each setting. Model rows average the four
prompting modes; the bottom block averages the six open-weight models
per mode. Unanswered or unparsable items count as wrong.}
\label{tab:photo_flips}
\end{table*}

\section{What the Photo Can Show}
\label{sec:appendix-vision-probe}

\paragraph{Visibility of the deciding evidence.}
Of the foods that decide the \textsc{not\_recommend} Task~1
questions with a listed food, 43.1\% are visible in the dish photo,
and in 261 of these questions none is; the other
\textsc{not\_recommend} questions fail a nutrient limit, which no
photo shows. On the 521 questions, photo-only accuracy is 68.4\% when
a deciding food is visible and 55.8\% when none is, whereas
recipe-only accuracy on the same two groups is 76.2\% and 79.6\%.

\paragraph{Vision probe.}
To measure how much of the visible evidence models read, we ask each
model to list the ingredients and cooking methods of the dish in each
Task~1 photo, with and without the dish title, and score the answers
against the recipe's ingredient list and instructions
(Figure~\ref{fig:vision_probe}). From the photo alone, models recover
70.0\% of the ingredients the photo shows, but only 37.9\% of the
others and 55.9\% of the gold rationale ingredients. They name the
cooking method that defines the dish in 63.4\% of cases, against
32.4\% for the most common method, but the labels depend on
ingredients, not on how a dish is cooked, so this evidence does not
move them. The title raises every figure, by 11.2 points for visible
ingredients and 6.9 for the cooking method, so part of what models
report from a photo comes from the dish name; even with the title,
they miss about 38\% of the rationale ingredients.
Table~\ref{tab:vision_probe} gives the per-model values behind
Figure~\ref{fig:vision_probe}, together with ingredient precision and
cooking-method accuracy over five coarse groups.

\begin{table*}[!tp]
\centering
\small
\setlength{\tabcolsep}{5pt}
\begin{tabular}{l l c c c c c c}
\toprule
 &  & \multicolumn{4}{c}{\textbf{Ingredients}} & \multicolumn{2}{c}{\textbf{Cooking method}} \\
\cmidrule(lr){3-6} \cmidrule(lr){7-8}
\textbf{Model} & \textbf{Input} & \textbf{Vis.\ R} & \textbf{Hid.\ R} & \textbf{Rat.\ R} & \textbf{Prec.} & \textbf{Final} & \textbf{Group} \\
\midrule
\multicolumn{2}{l}{\textit{Majority class}} & -- & -- & -- & -- & 32.4 & 39.5 \\
\midrule
\multirow{2}{*}{Gemma-3-12B-IT} & photo & 68.2 & 38.6 & 53.8 & 60.1 & 63.3 & 68.9 \\
 & + title & 79.9 & 42.9 & 59.1 & 66.0 & 70.9 & 75.7 \\
\cmidrule(lr){1-8}
\multirow{2}{*}{Gemma-3-27B-IT} & photo & 71.3 & 43.0 & 57.8 & 58.5 & 63.6 & 69.3 \\
 & + title & 82.0 & 46.5 & 64.4 & 65.1 & 70.9 & 76.2 \\
\cmidrule(lr){1-8}
\multirow{2}{*}{Gemma-4-31B-IT} & photo & 72.1 & 43.7 & 60.1 & 64.5 & 63.6 & 68.2 \\
 & + title & 82.5 & 48.0 & 64.4 & 70.8 & 70.9 & 74.6 \\
\cmidrule(lr){1-8}
\multirow{2}{*}{Qwen3-VL-8B-Instruct} & photo & 67.4 & 23.7 & 49.0 & 62.4 & 62.6 & 68.6 \\
 & + title & 78.4 & 28.6 & 54.2 & 69.7 & 68.2 & 73.7 \\
\cmidrule(lr){1-8}
\multirow{2}{*}{Qwen3-VL-32B-Instruct} & photo & 71.4 & 42.7 & 60.4 & 60.1 & 62.9 & 67.4 \\
 & + title & 83.5 & 49.5 & 68.3 & 66.4 & 69.1 & 72.9 \\
\cmidrule(lr){1-8}
\multirow{2}{*}{Qwen3.5-27B} & photo & 69.7 & 35.8 & 54.4 & 62.5 & 64.7 & 70.5 \\
 & + title & 81.2 & 42.4 & 61.0 & 69.8 & 71.9 & 76.4 \\
\midrule
\multirow{2}{*}{\textit{Open-weight mean}} & photo & 70.0 & 37.9 & 55.9 & 61.3 & 63.4 & 68.8 \\
 & + title & 81.2 & 43.0 & 61.9 & 68.0 & 70.3 & 74.9 \\
\midrule
\multirow{2}{*}{GPT-5.4} & photo & 73.3 & 41.0 & 56.0 & 62.9 & 67.9 & 72.2 \\
 & + title & 86.0 & 49.2 & 64.6 & 69.4 & 76.0 & 79.6 \\
\cmidrule(lr){1-8}
\multirow{2}{*}{Claude Sonnet 4.6} & photo & 77.6 & 56.8 & 65.2 & 59.0 & 67.9 & 72.6 \\
 & + title & 86.3 & 65.2 & 73.0 & 62.7 & 75.2 & 79.7 \\
\bottomrule
\end{tabular}
\caption{\textbf{What models read from the dish photo}: per-model
values behind Figure~\ref{fig:vision_probe}, from the photo alone or
with the dish title (+ title, the P input of
Table~\ref{tab:input_settings}). Ingredient recall over the
ingredients the photo shows (\textbf{Vis.\ R}), those it does not
(\textbf{Hid.\ R}; e.g., salt, oil, dried spices), and the gold
rationale ingredients (\textbf{Rat.\ R}); \textbf{Prec.}: share of
listed ingredients that are in the recipe. Ingredients and cooking
methods are extracted from the recipe text by GPT-5.4, and visibility
is judged from the photo by GPT-5.6-terra. \textbf{Final}: accuracy on
the cooking method that defines the finished dish (11 classes);
\textbf{Group}: the same over 5 coarse groups. Scored on the 1{,}164
recipes whose photo shows the dish.}
\label{tab:vision_probe}
\end{table*}

\section{Related Work}
\label{app:related_work}
\paragraph{Food recognition and recipe understanding.}
Food AI benchmarks have historically focused on recognizing dishes and structuring recipe content.
Food-101 established large-scale visual dish classification \cite{bossard2014food},
while Recipe1M+ paired recipes with food images for cross-modal retrieval and representation learning \cite{marin2021recipe1m+}.
RecipeQA extended food understanding to procedural reasoning over recipe text and images \cite{yagcioglu2018recipeqa},
and TASTEset introduced structured extraction of ingredients, quantities, cooking processes, and ingredient properties from recipes \cite{wroblewska2022tasteset}.
Generation-oriented corpora such as RecipeNLG further enabled controlled recipe synthesis at scale \cite{bien2020recipenlg}.
These resources provide the visual and linguistic foundations for food understanding,
but they remain descriptive: they identify what a dish is, what it contains, or how it is prepared,
without evaluating whether the dish is appropriate for a specific health condition.

\paragraph{Nutrition estimation and dietary-domain benchmarks.}
A second wave of benchmarks shifts from food recognition toward nutrient estimation and broader dietary reasoning.
Nutrition5k provides visual and nutritional annotations for real dishes \cite{thames2021nutrition5k},
and NutriBench evaluates large language models on calorie and macronutrient estimation from natural-language meal descriptions,
including a downstream blood-glucose simulation for type-1 diabetes \cite{hua2024nutribench}.
Recent multimodal benchmarks such as the January Food Benchmark and DiningBench broaden the setting to ingredient reasoning, nutrition estimation, and dietary-domain VQA \cite{hosseinian2025january, jin2026diningbench},
while FoodieQA curates multimodal Chinese food-culture QA but explicitly excludes nutrition and clinical content \cite{li2024foodieqa}.
These benchmarks make food AI more nutritionally grounded,
yet their targets remain primarily descriptive calories, macronutrients, ingredients, or food attributes rather than evaluating whether those food properties imply condition-specific dietary suitability.

\paragraph{Personalized nutrition and food-domain LLMs.}
A parallel line of work delivers system contributions: LLM-based pipelines that issue dietary advice.
ChatDiet couples user data with an LLM front-end for personalized nutrition counseling \cite{yang2024chatdiet};
NutriGen generates calorie-bounded meal plans under user preferences \cite{khamesian2025nutrigen};
HealthGenie grafts a knowledge graph onto an LLM for chronic-condition-aware recommendation \cite{gao2025healthgenie};
and NutriVision integrates vision with conversational AI for smart-healthcare diet management \cite{veeramreddy2024nutrivision}.
Constraint-aware approaches further survey ingredient substitution for chronic conditions \cite{kim2024survey}.
Food-specialized models adapt LLMs and vision--language models to culinary reasoning and dietetic knowledge:
LLaVA-Chef targets multimodal recipe generation \cite{mohbat2024llava},
and FoodSky reports strong performance on chef and dietetic examinations \cite{zhou2025foodsky}.
These contributions highlight the practical value of LLMs for nutrition support,
but each is evaluated on its own task suite,
making cross-system comparison and safety claims difficult to verify against a common standard.

\paragraph{Health-aware nutritional reasoning.}
The most closely related benchmark is NGQA \cite{zhang2025ngqa},
which formulates personalized nutritional assessment as graph question answering over user health profiles, dietary habits, foods, ingredients, and nutrient properties.
Built from NHANES and FNDDS records,
NGQA evaluates food suitability through binary classification,
identifies contributing nutrient properties through multi-label classification,
and assesses explanatory reasoning through text generation.
It therefore establishes an important precedent for evaluating both health-conditioned dietary decisions and their supporting explanations.
However, NGQA represents food and health information through structured graph relations and predefined nutrient tags,
rather than directly evaluating decisions from complete recipe records and dish photographs.
FAM-Bench complements this setting by assessing suitability from recipe-level evidence,
evaluating rationales against condition-relevant ingredients,
and introducing comparative ranking and cross-task consistency across candidate dishes.

Related work also investigates constrained question answering over food knowledge graphs \cite{chen2021personalized},
clinical health-aware text generation \cite{feng2023chard},
and knowledge-graph grounding for food-safety reasoning \cite{an2026knowledge}.
In the Chinese-medicine setting,
CMExam and TCMBench probe traditional-medicine knowledge through exam-style MCQs \cite{liu2023cmexam, yue2024tcmbench},
while two recent retrieval-augmented studies target \emph{medicine--food homology}:~\citet{gong2024integrating} integrate the TCM ``one root of medicine and food'' principle into dietary recommendations,
and Sha et al.~\cite{sha2025leveraging} report a validation study on TCM-styled cohorts.
Earlier work surveys TCM diet from a clinical-nutrition standpoint \cite{zhang2021diet}.
These studies investigate health-aware nutritional reasoning through structured knowledge representations, specialized question-answering tasks, and recommendation-oriented systems, but do not directly evaluate recipe-grounded dietary decisions across both individual and comparative dish settings.

\paragraph{Safety- and risk-aware evaluation of medical LLMs.}
A growing body of work shifts medical-LLM evaluation from accuracy alone toward safety and harm-awareness.
Med-PaLM and the MultiMedQA suite introduced multi-axis human evaluation including harm, factuality, and bias \cite{singhal2023multimedqa},
and MedHELM aggregates 35 benchmarks across 121 clinical tasks under an LLM-jury whose scores correlate with clinician judgment \cite{bedi2025medhelm}.
Hallucination benchmarks such as Med-HALT \cite{pal2023medhalt} and atomic-fact methods like FActScore \cite{min2023factscore} target factuality failures,
while the recent RCT of Li et al.~\cite{li2025whenllms} documents real-world failures of LLM-augmented nutrition chatbots and explicitly calls for risk-aware intrinsic evaluation.
Adjacent to safety,
``Cooking Up Risks'' stress-tests LLMs against adversarial food-safety prompts,
targeting misuse rather than well-intentioned but clinically risky advice \cite{luo2026cookingup}.
Existing food-safety benchmarks themselves remain centered on contamination, storage, and harmful preparation
\cite{jacxsens2010food, le2015benchmarking, bryan1992hazard, muncke2025health, pekmezci2025health}.
Despite this momentum,
nutrition and chronic-condition-aware dietary advice remain underrepresented in safety-oriented medical-LLM evaluation:
the existing literature offers strong methodological precedent but limited coverage of the dietary domain.

\paragraph{Reasoning and retrieval for health-aware generation.}
Two prompting strategies dominate recent work on knowledge-intensive medical generation.
Chain-of-Thought prompting \cite{wei2022cot, kojima2022zeroshot} improves multi-step reasoning,
and Medprompt demonstrates large gains on clinical MCQA through CoT-style prompting and ensembling \cite{nori2023medprompt}.
Retrieval-Augmented Generation grounds outputs in external knowledge \cite{lewis2020rag},
and medical extensions such as MedRAG / MIRAGE show that retrieval over curated biomedical corpora substantially reduces hallucination \cite{xiong2024medrag}.
In the food domain,
retrieval has been applied to TCM medicine--food homology \cite{gong2024integrating, sha2025leveraging}
and to knowledge-graph grounding for dietary guidance \cite{gao2025healthgenie, an2026knowledge}.
While individual systems incorporate CoT or RAG,
a systematic comparison of these strategies on dietary recommendation across heterogeneous health conditions has not, to our knowledge, been reported.

\end{document}